\documentclass[letterpaper, 9 pt, conference]{ieeeconf}  

\IEEEoverridecommandlockouts                              

\usepackage{cite}
\usepackage{amsmath}
\usepackage{amsfonts}
\usepackage{amssymb}
\usepackage{graphicx}
\usepackage{enumerate}
\usepackage{subfigure}
\usepackage{mathrsfs}
\usepackage{braket}
\usepackage{algorithm}
\usepackage{algpseudocode}
\usepackage{tikz}
\usetikzlibrary{arrows, shapes, snakes, automata, backgrounds, petri, matrix, calc, shapes}
\usepackage{hyperref}
\title{\LARGE \bf
Robotic Playing for Hierarchical Complex Skill Learning
}

\author{Simon Hangl, Emre Ugur, Sandor Szedmak and Justus Piater$^{1}$
\thanks{$^{1}$Department of Computer Science,
        University of Innsbruck, 6020 Innsbruck, Austria
        {\tt\small first.last@uibk.ac.at}}%
}

\tikzset{
  treenode/.style = {shape=rectangle, rounded corners,
                     draw, anchor=center,
                     text width=5em, align=center,
                     top color=white, bottom color=blue!20,
                     inner sep=1ex},
  decision/.style = {treenode, diamond, inner sep=0pt},
  root/.style     = {treenode, bottom color=red!30},
  env/.style      = {treenode, font=\ttfamily\normalsize},
  finish/.style   = {root, bottom color=green!40},
  dummy/.style    = {}
}

\begin{document}

\maketitle
\thispagestyle{empty}
\pagestyle{empty}

\begin{abstract}
In complex manipulation scenarios (e.g.\ tasks requiring complex interaction of two hands
or in-hand manipulation), generalization is a hard problem. Current methods still
either require a substantial amount of (supervised) training data
and / or strong assumptions on both the environment and the task. In this paradigm, controllers
solving these tasks tend to be complex. We propose
a paradigm of maintaining simpler controllers solving the task
in a small number of specific situations. In order to
generalize to novel situations, the robot transforms the environment
from novel situations into a situation where
the solution of the task is already known. Our solution
to this problem is to play with objects and use previously
trained skills (basis skills). These skills can either be used for estimating
or for changing the current state of the environment and are organized in skill hierarchies.
The approach is evaluated in
complex pick-and-place
scenarios that involve complex manipulation. We further show that these
skills can be learned by autonomous playing.
\end{abstract}
\section{Introduction}
Complex object manipulation in uncontrolled environments is a hard and not yet completely solved
problem in robotics.
One of the major issues in this context is the problem of generalizing motor skills \cite{Argall:2009:SRL:1523530.1524008, Smith2012, mulling2013learning, Hangl-2015-ICAR}.
Much of it incorporates a paradigm where the aim is to adapt the controller itself to
the changing environments. This increases the complexity of
the manipulation controller, as it should deal with a wide range of different situations.

We propose to combine simpler and previously-learned skills in order to achieve more complex tasks.
The aim is to exploit simple skills to transfer the environment into a state where
simple controllers can achieve the desired complex task. This allows the complexity of
the controllers to be reduced, as they do not have to deal with generalization.

Humans use similar behavioural patterns e.g.\ in sports such as golf. The player always tries
to stand in the same position relative to the ball instead of adapting the swing
itself in order to hit the ball from another position. Therefore, the player is able to execute the
same (or very similar), previously-learned trajectories. This can highly reduce the training cost 
by constraining the search space. We emphasize that in most approaches in robotics, the robot
would have to adapt the swing in order to hit the ball from many different positions.
A similar strategy seems to be exploited by human infants.
Piaget observed similar patterns in infant playing at the age between 8 and 12 months \cite{piaget}.
This stage in the life of infants
is called the \emph{coordination of secondary schemata} and Piaget calls it the stage of \emph{first actually
intelligent behaviours}. Infants use previously-learned skills to bring the objects into a state where they can
perform an intended action (e.g.\ kicking an obstacle out of the way to grasp an object; pulling a string attached
to an object to bring it within reach). An important property of this stage is that they do not predict the effects
of these actions directly, but rather learn to compose previously-known skills to achieve a specific task. They do not have
an understanding of what the effect of a manipulation is. However, they know that a composition of certain
skills leads to a successful manipulation.
\begin{figure}[t!]
 \centering
	 \includegraphics[width=0.4\textwidth]{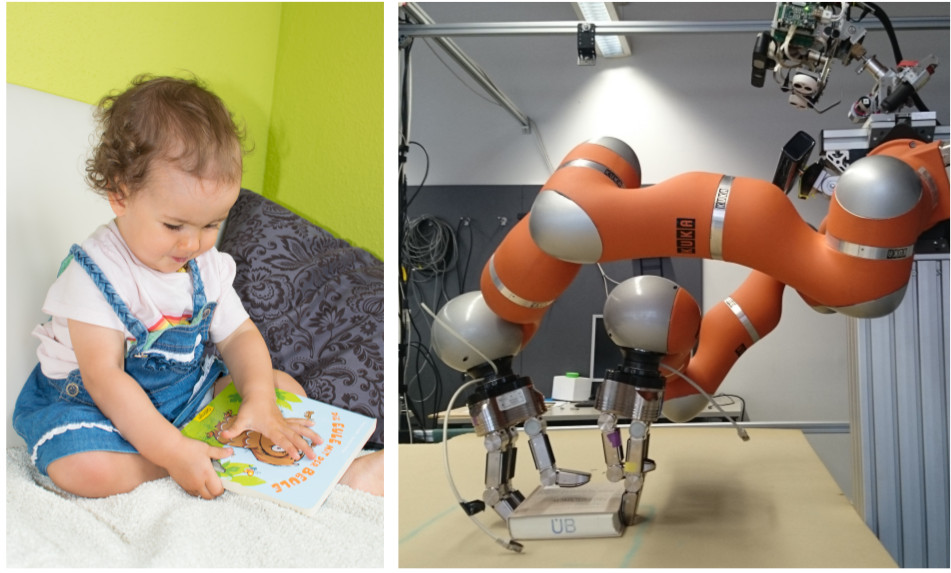}
  \caption{Book manipulation: Infant vs. autonomous robot}
  \label{fig:trajGenb}
\end{figure}
In this paper we propose a method that follows a similar paradigm. The robot holds a set of \emph{preparatory skills}.
These are used to bring the system from an arbitrary state to a state where a desired complex skill can be executed
with limited generalization demands. Let $\mathbf{s}$ be the complete (and unknown) physical state of the system
the robot is located in. We use a set $S$ of \emph{sensing actions} $S_i \in S$ and the haptic sensor data $\mathbf{t}_{S_i}\left(
\mathbf{s}\right) \propto p\left( \mathbf{t}_{S_i} \right | \mathbf{s})$ collected during the execution of $S_i$ to gather task-relevant information from the environment. We use a classifier to estimate discrete sensing action-dependent state labels
$E_{S_ij} \propto p \left( E_{S_i} = E_{S_ij} | \mathbf{t}_{S_i} \right)$. In the remainder of the paper we will
refer to these estimated labels $E_{S_ij}$ as the \emph{perceptual state} (given the sensing action $S_i$). These labels
can be predefined by supervision or generated from experience. Given the observed state $E_{S_ij}$ the robot will pick an
appropriate preparatory skill. After preparation, the complex skill is executed in order to achieve a desired task.
For clarity we will illustrate the single components in a tabletop book grasping scenario (Fig. \ref{fig:trajGenb}). In this
case the robot cannot perform conventional grasping strategies because it cannot get the fingers below book.
The \emph{complex skill} involves the coordination of two hands, where one hand prevents the book from sliding while
the other presses against the binding of the book and tries to lift it. The second hand performs an
in-hand manipulation to position a finger underneath the book in order to finally grasp it (Fig.~\ref{fig:trajGenb}). The relevant \emph{perceptual state}, namely the robot-relative orientation of the book,
can be determined by a sliding action (\emph{sensing action}) along the book surface. The complex behaviour is
shown for one specific orientation. Pushing controllers might be handy to prepare this orientation from
an arbitrary orientation. Therefore, pushing controllers are good \emph{preparatory skills} for this task.

The learning problem solved in this paper is how to autonomously learn to generalize complex skills
shown by kinesthetic teaching (or hard-coding) within the paradigm described above.
This involves the selection of the
best sensing action and the best preparatory skill
given a specific perceptual state in a so-called \emph{playing phase}. We do this by using
a reinforcement learning method called \emph{projective simulation} (PS \cite{Briegel2012}) which is well
suited for this type of learning problems.
Further, we show how the sensor data gathered during the sensing action can be mapped to
the discrete perceptual state by a data-driven classifier called \emph{Maximum Margin Regression}
(MMR \cite{szs2005lo}). In the playing phase the data required to initialize the
classifier is generated autonomously as well. We also show how the robot can create skill hierarchies
by adding complex skills to its repertoire of preparatory skills. In this way, the robot
can learn increasingly more complex tasks over time.
\section{Related Work}
Belief-space planning where the systems state is partially observed by sensors is similar in nature.
In most belief-space planning methods a control policy is trained and at each time step the next commands are predicted \cite{beliefspace1, beliefspace2}.
These commands are given in the action space of the robot, while we select complete
controllers. This causes a significant reduction of the learning complexity.
Such macro actions were used in a
navigation task to reduce the dimensionality \cite{beliefspace3}.
However, these macro actions were limited to the navigation domain (e.g.\ relocation primitives).
In a manipulation scenario, this method would require pre-defined primitives, which is hard
to achieve generally.
Other related work uses haptic feedback to derive information about the environment. Robots can
learn the meaning of haptic adjectives that were
previously assigned to a set of objects by interacting (tap, squeeze, slide, $\dots$) with them \cite{Chu_usingrobotic}.
Similarly, interaction primitives were used to classify objects by using haptic feedback clustered with K-means \cite{objectidenthaptic}.
Similar in spirit, these methods do only deal with the estimation of properties and not with manipulation.
Associative skill memories on the other hand \cite{associativeskillmem} assign typical task-specific force patterns
to manipulations. The patterns are used to predict the success of manipulation during execution.
Therefore the robot can react in time and change the trajectory accordingly.
Jain et. al. transferred haptic time series collected from stereotypical tasks performed by humans
to robotic manipulation \cite{datadrivenforcemanipulation}. The data was used categorize objects
or detect anomalies during the execution.
Manipulation primitives have been proposed in order to estimate object poses
and further afforded actions \cite{6100911}. Primitives are composed in order
to transfer the object to a pose in which a task can be executed. Even though
using similar ideas, our approach is not limited to object poses as state space.
Vigorito et. al. \cite{Vigorito2010IntrinsicallyMH} 
predicted manipulation effects (in contrast to our method) and composed skills by planning in state space
by optimizing intrinsic reward.
Another class of competitors are logic-based planning systems, e.g.\ STRIPS \cite{strips}, where
provably-correct plans are derived to achieve a certain goal by matching pre- and post-conditions
of action primitives. This requires a higher level of abstraction and often prior knowledge (e.g.,
how abstract symbols are created from real-world data).
In contrast, our method does not predict outcomes of actions,
but learns successful sequences of actions in an open loop.
Open-loop planners for grasping \cite{pushgrasp, pushforgrasping} rearrange the objects in clutter in order
to perform simpler grasps. However, this method is restricted to the grasping domain.
Applications of deep learning to robotics (Levine et. al. \cite{levine2016learning}) are interesting in this context as they require
a high autonomy because of the huge data demand. Large scale autonomous experiments were performed to train
a CNN for grasp success prediction, which is used to servo the robot in a closed loop. In contrast to
our work they do not need to design internal representations but require a huge amount of data.
\section{System architecture}
\label{sec:architecture}
The method comprises two interrelated pathways,
the \emph{execution pathway} and the \emph{playing pathway}.
The \emph{execution pathway} is used to execute a complex skill, i.e.\ to execute the sensing action,
estimate the perceptual state from haptic data, perform the preparatory action and finally
execute the complex skill. Initially, the system does not know which sensing action and which
preparatory skills are required to achieve a certain task. It also has no information about
what haptic feedback corresponds to which predefined (or automatically generated) discrete perceptual states.
The \emph{playing pathway} is used to acquire this information by playing with the object.
In order to train a novel skill, the robot needs to gather haptic information about the
perceptual states it might observe (e.g.\ the rotation of a book or information whether a
box is opened or closed) and explores the environment with its sensing actions.
Each action $S_i$ is assumed to leave the perceptual states $E_{S_ij}$ unchanged
(e.g., they do not change the rotation of a book) and can therefore be performed multiple times.
This is important in order to create a haptic database for each state $E_{S_ij}$.
After creating this database the system is trained to \emph{select a sensing action}
and to \emph{pick the preparation skill} that ensures the successful execution
of the novel skill given an estimated perceptual state.
This is achieved by a reinforcement learning method called
\emph{projective simulation} (PS) \cite{Briegel2012}. In each roll-out (skill execution, reward collection and model update)
the execution pathway is performed and the reward is measured
until the success rate reaches a certain threshold.
When the success rate for the novel skill is high enough, it is added to the set of preparatory skills.
This way, the construction of skill hierarchies is possible, as a complex skill can be used as a preparatory action for another
complex skill. For example, placing a book on a shelf requires grasping it first. If the
robot only knows how to push objects, it will not be able to perform the complex placing action. However,
as soon as it has learned how to properly grasp a book, it can do the placement by using
grasping as a preparatory skill.
\subsection{Execution pathway}
\label{sec:execpathway}
\begin{figure*}[t!]
  \subfigure[\label{fig:overview}{\footnotesize{
  Overview of \emph{playing pathway}: Training of the haptic properties, selection of the best sensing actions and preparatory skills by reinforcement learning}}]{
\scalebox{0.6} {
\begin{tikzpicture}
	\node[root] (w1) {Teach novel complex skill};
	\node[env] (w2) [right of = w1, xshift = 7em] {Create sensing database};
	\node[dummy] (d1) [below of = w2, yshift = -1em] {};
	\node[env] (w3) [below of = w2, yshift = -5em] {Init / Update ECM};
	\node[env] (w4) [below of = w3, yshift = -3em] {Execute Complex Skill};
	\node[decision] (w5) [below of = w4, yshift = -5em] {Reached Confidence Threshold?};
	\node[env] (w6) [left of = w5, xshift = -7em] {Add to ECM as Prep Action};
	\node[dummy] (d2) [right of = w5, xshift = 4em] {};
	\node[dummy] (d3) [right of = d1, xshift = 4em] {};
	
	\path[->] (w1) edge node {} (w2);
	\path[->] (w2) edge node {} (w3);
	\path[->] (w3) edge node {} (w4);
	\path[->] (w4) edge node {} (w5);
	\path[->] (w5.west) edge node {} (w6.east);
	\path[->] (w6) edge node {} (w1);
	\path[-] (w5.east) edge node {} (d2);
	\path[-] (d2.north) edge node {} (d3);
	\path[->] (d3.west) edge node {} (d1.east);
\end{tikzpicture} 
}
}\hfill
\subfigure[\label{fig:architecture}{\footnotesize{Schematic sketch of the \emph{execution pathway}.
The robot senses the task-relevant perceptual state by haptic exploration and
classifies the measured data. The estimated haptic class is mapped to the preparatory action by the relation learning
module. After preparing the environment with the preparatory action, the complex skill can be executed.}}]{
\includegraphics[width=0.7\textwidth]{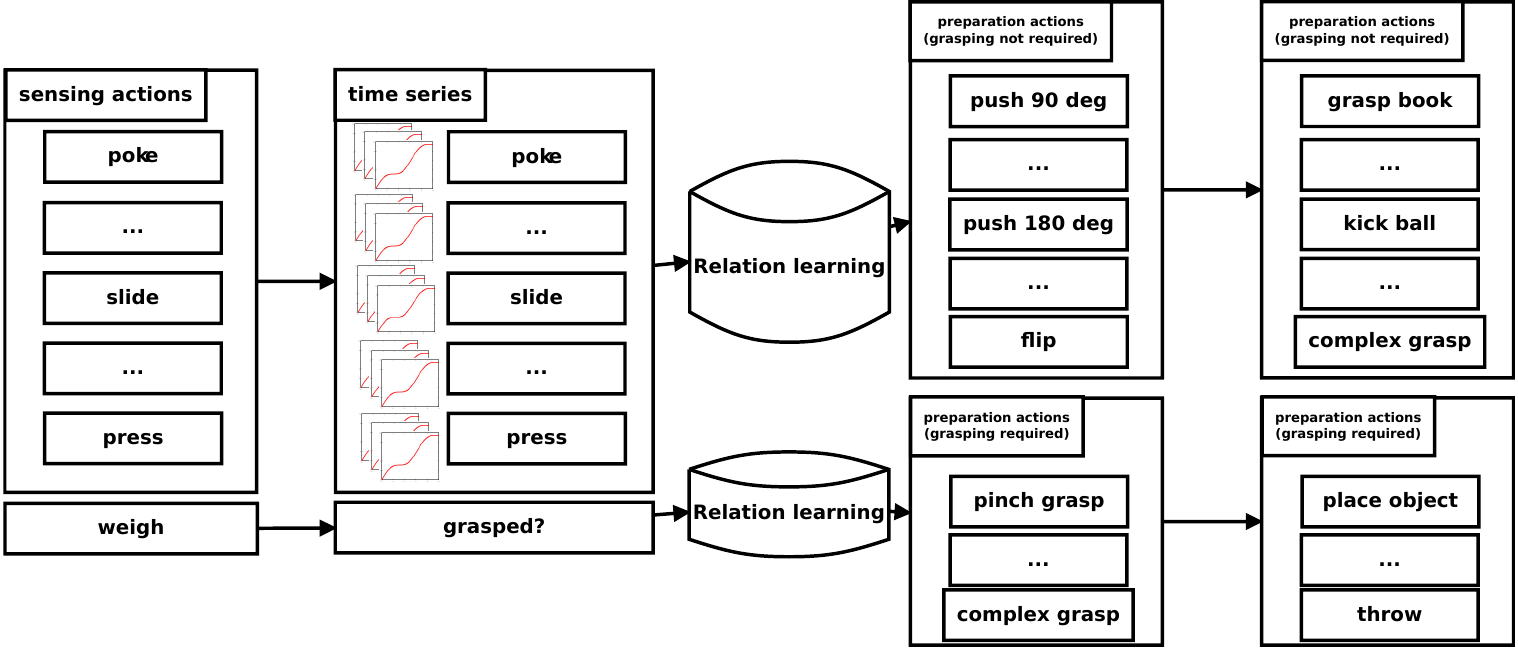}}
\caption{Schematic sketch of the two parts of the proposed method: the \emph{playing pathway} (left) and the \emph{execution pathway.}}
\end{figure*}
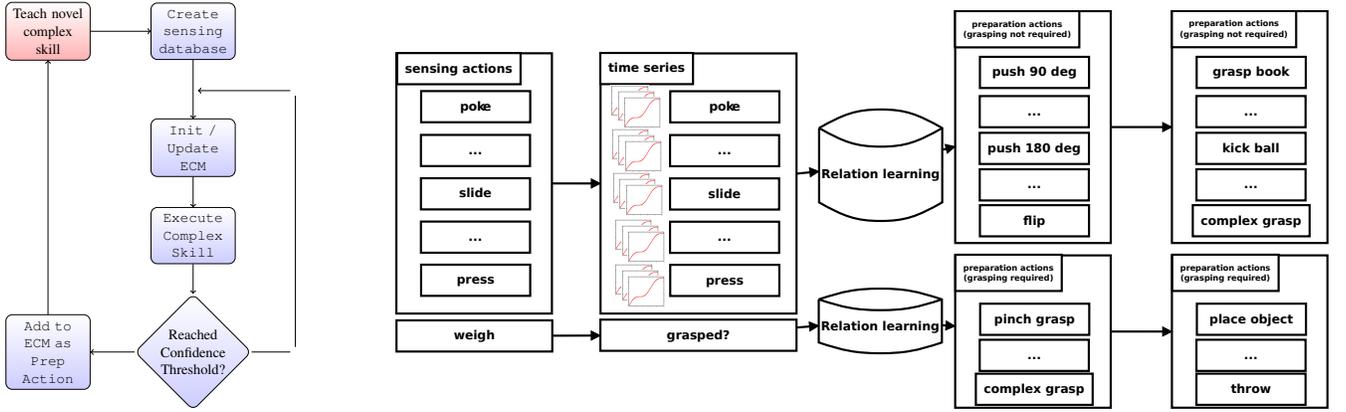

In order to execute a novel complex skill that was trained by playing,
the following steps are executed (Fig. \ref{fig:architecture}):
\begin{itemize}
	\item \emph{Select a sensing action} in order to collect data for perceptual state estimation.
	\item \emph{Perform the sensing action} and measure haptic data.
	\item \emph{Estimate the perceptual state} by classifying the haptic data.
	\item \emph{Select and execute a preparatory skill} to transform the environment into a state in which the complex action can be executed successfully.
	\item \emph{Execute the complex skill}, e.g.\ by replay of trained trajectories or execution of hard-coded controllers.
\end{itemize}
\begin{figure}[t!]
 \centering
	 \includegraphics[width=0.49\textwidth]{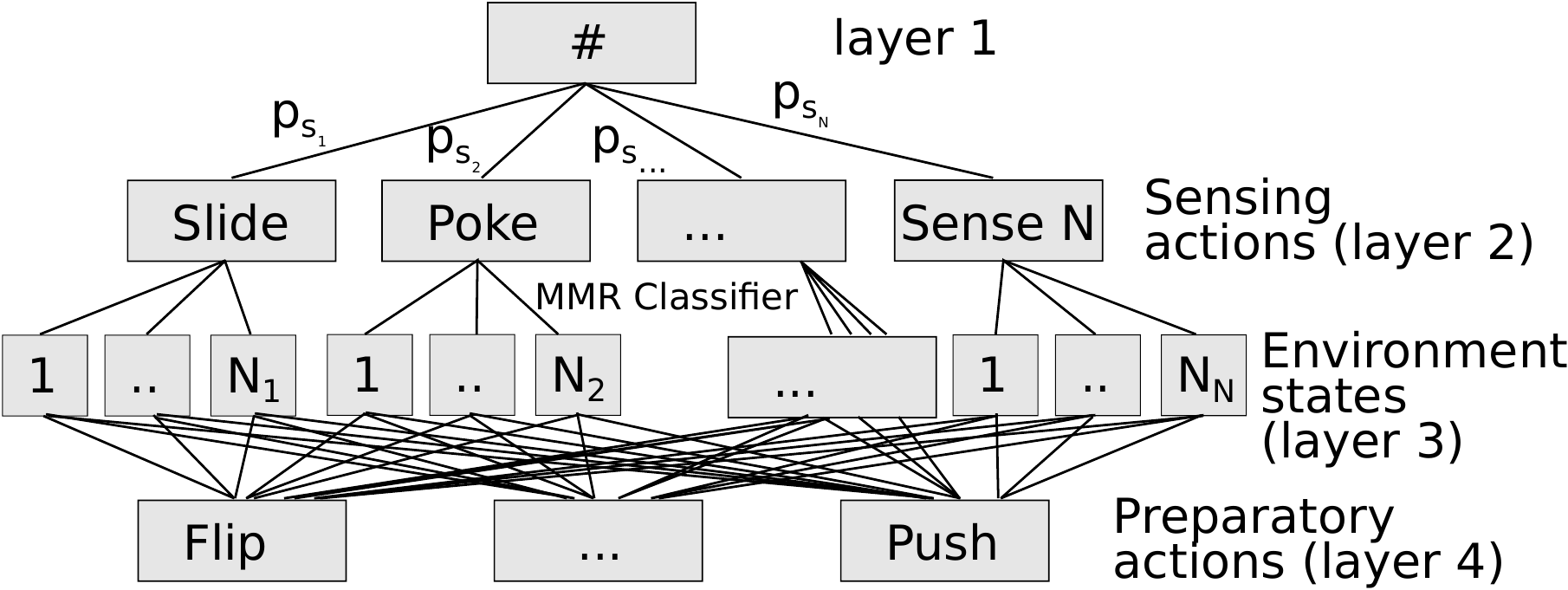}
  \caption{ECM in the hierarchical skill learning scenario for one complex skill.}
  \label{fig:hapticps}
\end{figure}
\subsubsection{relational model for sensing / preparation pairs}
\label{sec:psstructure}
The relational model is the heart of the method and is used for two essential tasks: (1) selecting
the best sensing action given a desired task, and (2) selecting the correct preparatory action
given the estimated perceptual state. We use projective simulation (PS) for skill execution
and skill learning. PS is a reinforcement learning method that consists of an \emph{episodic and compositional memory} (ECM)
which is a network of so-called \emph{clips}, i.e., episodic memory fragments that include percepts and actions.
Each complex skill is encoded by an ECM with the novel, fixed, layered structure proposed in this work (Fig. \ref{fig:hapticps}).
A complex skill is executed by a \emph{random walk} between clips through the ECM.
A transition probability $p\left( c_j | c_i \right)$ is assigned to each pair of
clips $c_i, c_j$ that is connected in Fig. \ref{fig:hapticps} with
\begin{equation}
	p\left( c_j | c_i \right) = \frac{h\left( c_i, c_j \right)}{\sum_k h\left(c_i, c_k \right)}
	\label{equ:randwalk}
\end{equation}
where $h(c_i, c_j)$ is called the \emph{transition weight}. In the first step (transition
from layer 1 to layer 2), a sensing action $S_i$ is selected and executed. A time series
$t_{S_i} = \{ (t_{{S_i}k}, \mathbf{F}_{{S_i}k}, \mathbf{T}_{{S_i}k}, \mathbf{P}_{{S_i}k}) \}$ of haptic data is measured,
where $t_{{S_i}k}$ is the $k$-th time step of the observed time series, $\mathbf{F}_{{S_i}k}$ is the
Cartesian force, $\mathbf{T}_{{S_i}k}$ is the torque and $\mathbf{P}_{{S_i}k}$ is the Cartesian end-effector position.
From this data, the perceptual state is estimated using
a multi-class classifier, namely \emph{maximum margin regression} (MMR)
\cite{szs2005lo}. The multi-dimensional input is simply a concatenated vector
$\mathbf{t}_{S_i} = \left( \mathbf{F}_{{S_i}0}, \mathbf{T}_{{S_i}0}, \mathbf{P}_{{S_i}0}, \dots, \mathbf{F}_{{S_i}T}, \mathbf{T}_{{S_i}T}, \mathbf{P}_{{S_i}T}\right)$. The output is an $N_{S_i}$-tuple, where $N_{S_i}$ is the number of classes for
sensing action $S_i$. If the time series belongs to class $E_{S_ij}$, only the $j$-th entry
is nonzero. MMR is simple to use, the code is available online\footnote{\url{https://iis.uibk.ac.at/software/mmr\_mmmvr}},
and it was shown to perform well on different complex tasks \cite{Mustafa-2015-3DV}.
In this step, the walk is not random but a transition from clip $S_i$ to the $j$-th child is performed
according to the classifier output. The transition between layers
3 and 4 and therefore the selection of the preparatory skill is again random
according to equation \ref{equ:randwalk}, and the corresponding preparatory skill is executed.
Finally, the robot performs the desired complex skill.

Complex skills that require an object to be grasped are treated separately.
The only such sensing action is \emph{weighing}. If the external force
$F = 0$, the object is not grasped yet, and only preparatory actions
that result in a grasp are considered. Otherwise (i.e., the object
is already grasped) the novel complex skill can be executed directly.
If the complex skill does not require a grasp, then only the preparatory actions that do not result
in a grasp are considered (Fig. \ref{fig:architecture}).
\subsection{Playing pathway}
\label{sec:play}
A novel complex skill (e.g.\ shown by hard-coding a controller, kinesthetic teaching or
providing more complex controllers with some limited amount of generalization) is learned
by playing with the object (Fig. \ref{fig:overview}). The purpose is to learn the best
transition weights $h(c_m, c_n)$ and to create a haptic database for state estimation.
\subsubsection{Creation of the haptic database}
The haptic database is required to initialize the MMR classifier for state estimation.
The perceptual state is a discrete class that determines some aspect of the environment.
It can, but does not have to, have a semantic meaning (e.g.\ whether a box is open or not, or
the pose of an object). The haptic time series are labelled with the corresponding
state $E_{S_ij}$ they were measured in for all sensing actions $S_i$.
In order to do so, each perceptual state $E_{S_ij}$ has to be created. This can be done
in a supervised or unsupervised manner. In other words, either a supervisor prepares the state
(e.g.\ a human shows the robot a box when it is open or closed) or the robot tries to prepare
the states by itself. If no supervisor is available to play with the robot, it uses the preparatory
skills to change the state of the environment (e.g.\ rotating a book by 90 degrees in
each iteration). After each execution the system is assumed to be in a novel state
and the haptic time series are collected.

Not all sensing actions have the same discriminative power for every
task / object (e.g., while a sliding action is good for deciding the
orientation of a book, a poking action is not helpful in that case).
Therefore, for each complex skill and object class we assign a \emph{discrimination score}
$D_i$ to each sensing action $S_i$. It is computed by performing
cross-validation for the time series classifier.
With an average success rate of $s_i$, the discrimination score $D_i$ is given by
\begin{equation}
	D_i = \textrm{exp}(\alpha s_i)
	\label{equ:discrim}
\end{equation}
with a fixed \emph{stretch factor} $\alpha$. The higher the stretch factor, the
more strongly slight differences in the average success rate influence the discrimination score.
The score $D_i$ determines how well the sensing action $S_i$ can distinguish its
states $E_{S_ij}$.
\begin{figure}[t!]
\centering
  \scalebox{0.8} {
\begin{tikzpicture}[every node/.style = {shape=rectangle, rounded corners},>=stealth',bend angle=45, auto]

  \tikzstyle{place}=[circle,thick,draw=blue!75,fill=blue!20,minimum size=6mm]
  \tikzstyle{red place}=[place,draw=red!75,fill=red!20]
  \tikzstyle{transition}=[rectangle,thick,draw=black!75,
  			  fill=black!20,minimum size=4mm]

  \tikzstyle{every label}=[red]

  \begin{scope}
                      
	\node [place,tokens=1] (w1)                                                {};
    \node [place]          (c1) [below of=w1,xshift=-25mm]  {Slide};
    
    \node [transition]          (s1) [below of=c1,xshift=-25mm]  {bottom};
    \node [transition]          (s2) [below of=c1,xshift=-10mm]  {binding};
    \node [transition]          (s3) [below of=c1,xshift=5mm]  {open};
    \node [transition]          (s4) [below of=c1,xshift=16mm]  {top};
    
    \node [place]          (c2) [below of=w1]  {Poke};
    
	\node [transition]          (s5) [below of=c2,xshift=-2mm]  {1};
    \node [transition]          (s6) [below of=c2,xshift=3mm]  {2};
    \node [transition]          (s7) [below of=c2,xshift=8mm]  {3};
    \node [transition]          (s8) [below of=c2,xshift=13mm]  {4};
    
    \node [place]          (c3) [below of=w1,xshift=25mm]  {Press};
    
    \node [transition]          (sc1) [below of=c3,xshift=-5mm]  {1};
    \node [transition]          (sc2) [below of=c3,xshift=0mm]  {2};
    \node [transition]          (sc3) [below of=c3,xshift=5mm]  {3};
    \node [transition]          (sc4) [below of=c3,xshift=10mm]  {4};
    
    \node [transition]          (p1) [below of=c2, xshift=-50mm, yshift=-20mm, minimum size=6mm]  {push 90};
    \node [transition]          (p2) [below of=c2, xshift=-25mm, yshift=-20mm, minimum size=6mm]  {push 180};
    \node [transition]          (p3) [below of=c2, xshift=-5mm, yshift=-20mm, minimum size=6mm]  {push 270};
    \node [transition]          (p4) [below of=c2, xshift=10mm, yshift=-20mm, minimum size=6mm]  {flip};
    \node [transition]          (p5) [below of=c2, xshift=30mm, yshift=-20mm, minimum size=6mm]  {nothing};

	\path[->] (w1) edge [line width=0.6mm, red] node {} (c1);
	\path[->] (w1) edge node {} (c2);
	\path[->] (w1) edge node {} (c3);
	
	\path[->] (c1) edge node {} (s1);
	\path[->] (c1) edge node {} (s2);
	\path[->] (c1) edge node {} (s3);
	\path[->] (c1) edge node {} (s4);
	
	\path[->] (c2) edge node {} (s5);
	\path[->] (c2) edge node {} (s6);
	\path[->] (c2) edge node {} (s7);
	\path[->] (c2) edge node {} (s8);
	
	\path[->] (c3) edge node {} (sc1);
	\path[->] (c3) edge node {} (sc2);
	\path[->] (c3) edge node {} (sc3);
	\path[->] (c3) edge node {} (sc4);
	
	\path[->] (s1.south) edge node {} (p1);
	\path[->] (s1.south) edge node {} (p2);
	\path[->] (s1.south) edge node {} (p4);
	\path[->] (s1.south) edge node {} (p5);
	
	\path[->] (s2.south) edge node {} (p1);
	\path[->] (s2.south) edge node {} (p2);
	\path[->] (s2.south) edge node {} (p3);
	\path[->] (s2.south) edge node {} (p4);
	
	\path[->] (s3.south) edge node {} (p1);
	\path[->] (s3.south) edge node {} (p3);
	\path[->] (s3.south) edge node {} (p4);
	\path[->] (s3.south) edge node {} (p5);
	
	\path[->] (s4.south) edge node {} (p2);
	\path[->] (s4.south) edge node {} (p3);
	\path[->] (s4.south) edge node {} (p4);
	\path[->] (s4.south) edge node {} (p5);
	
	\path[->] (s5.south) edge node {} (p1);
	\path[->] (s5.south) edge node {} (p2);
	\path[->] (s5.south) edge node {} (p3);
	\path[->] (s5.south) edge node {} (p4);
	\path[->] (s5.south) edge node {} (p5);
	
	\path[->] (s6.south)  edge node {} (p1);
	\path[->] (s6.south)  edge node {} (p2);
	\path[->] (s6.south)  edge node {} (p3);
	\path[->] (s6.south)  edge node {} (p4);
	\path[->] (s6.south)  edge node {} (p5);
	
	\path[->] (s7.south)  edge node {} (p1);
	\path[->] (s7.south)  edge node {} (p2);
	\path[->] (s7.south)  edge node {} (p3);
	\path[->] (s7.south)  edge node {} (p4);
	\path[->] (s7.south)  edge node {} (p5);
	
	\path[->] (s8.south)  edge node {} (p1);
	\path[->] (s8.south)  edge node {} (p2);
	\path[->] (s8.south)  edge node {} (p3);
	\path[->] (s8.south)  edge node {} (p4);
	\path[->] (s8.south)  edge node {} (p5);
	
	\path[->] (sc1.south) edge node {} (p1);
	\path[->] (sc1.south) edge node {} (p2);
	\path[->] (sc1.south) edge node {} (p3);
	\path[->] (sc1.south) edge node {} (p4);
	\path[->] (sc1.south) edge node {} (p5);
	
	\path[->] (sc2.south) edge node {} (p1);
	\path[->] (sc2.south) edge node {} (p2);
	\path[->] (sc2.south) edge node {} (p3);
	\path[->] (sc2.south) edge node {} (p4);
	\path[->] (sc2.south) edge node {} (p5);
	
	\path[->] (sc3.south) edge node {} (p1);
	\path[->] (sc3.south) edge node {} (p2);
	\path[->] (sc3.south) edge node {} (p3);
	\path[->] (sc3.south) edge node {} (p4);
	\path[->] (sc3.south) edge node {} (p5);
	
	\path[->] (sc4.south) edge node {} (p1);
	\path[->] (sc4.south) edge node {} (p2);
	\path[->] (sc4.south) edge node {} (p3);
	\path[->] (sc4.south) edge node {} (p4);
	\path[->] (sc4.south) edge node {} (p5);
	
	\path[->] (s1) [line width=0.6mm, green, out = -90] edge node {} (p3);
	\path[->] (s2) [line width=0.6mm, blue, out = -90] edge node {} (p5);
	\path[->] (s3) [line width=0.6mm, red, out = -90] edge node {} (p2);
	\path[->] (s4) [line width=0.6mm, orange, out = -90] edge node {} (p1);
	
  \end{scope}
\end{tikzpicture}
}
\caption{Qualitative sketch of the episodic memory after learning how to grasp a book. Coloured lines
indicate a high probability of the transition. Semantic labels are assigned to perceptual states if available.}
\label{fig:pickQualRes}
\end{figure}
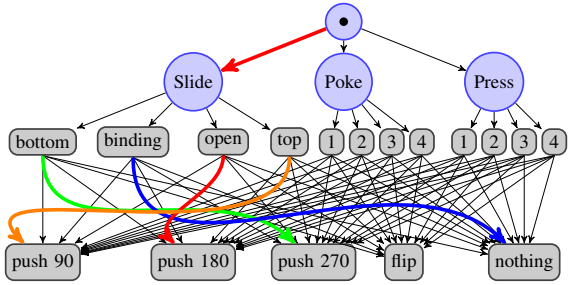
\subsubsection{Initialization of the ECM}
After the classifier is trained, the robot can start with the exploration of sensing and preparation actions.
The ECM has to be initialized, setting the transition weights $h(c_i, c_j)$.
Sensing actions that can discriminate well between their states should be preferred, and the discrimination
score $D_i$ can be used as an initial transition weight between the \#-clip (starting clip of each random walk) and clip $S_i$ with $p_{S_i} \propto h_{S_i} = D_i$.
The transition probabilities between layers 2 and 3 are given by the time series classifier,
where only the transition to the predicted state has nonzero probability.
The weights from layer 3 to layer 4 are initialized with the constant value $h_{\text{init}}$ (uniform distribution).
\subsubsection{Relation learning}
\begin{figure*}
\centering
  \subfigure[\label{fig:robotsetting}{}]{\includegraphics[height=0.18\linewidth]{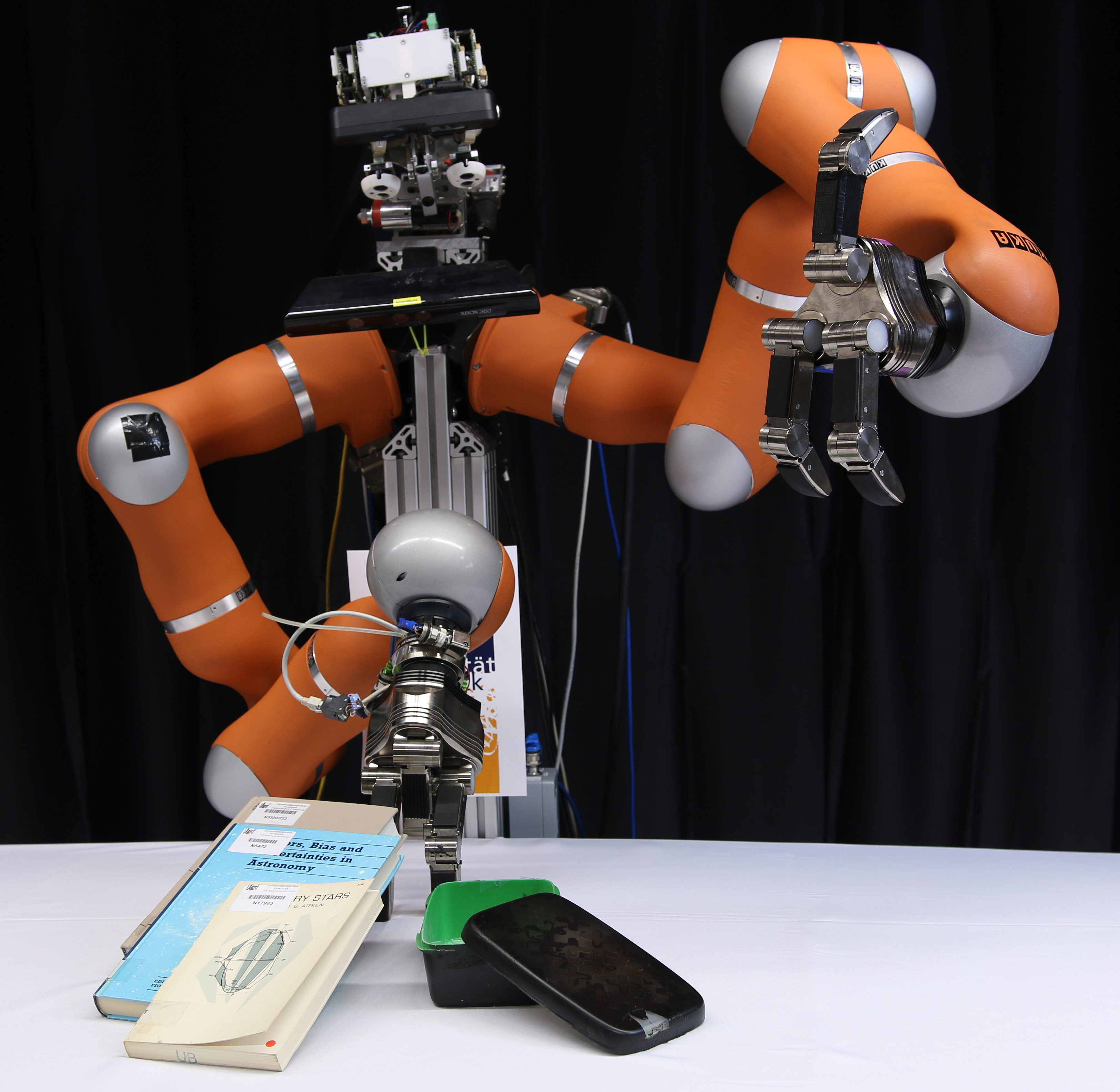}}
  \subfigure[\label{fig:pick1}{}]{\includegraphics[height=.18\linewidth]{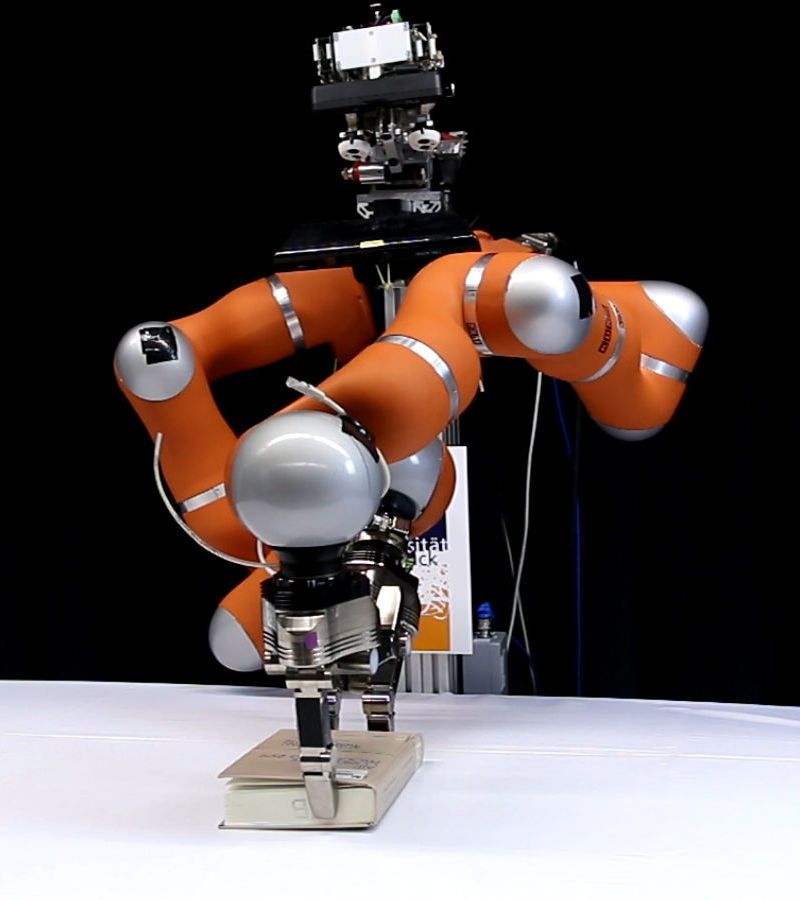}}
  \subfigure[\label{fig:pick2}{}]{\includegraphics[height=.18\linewidth]{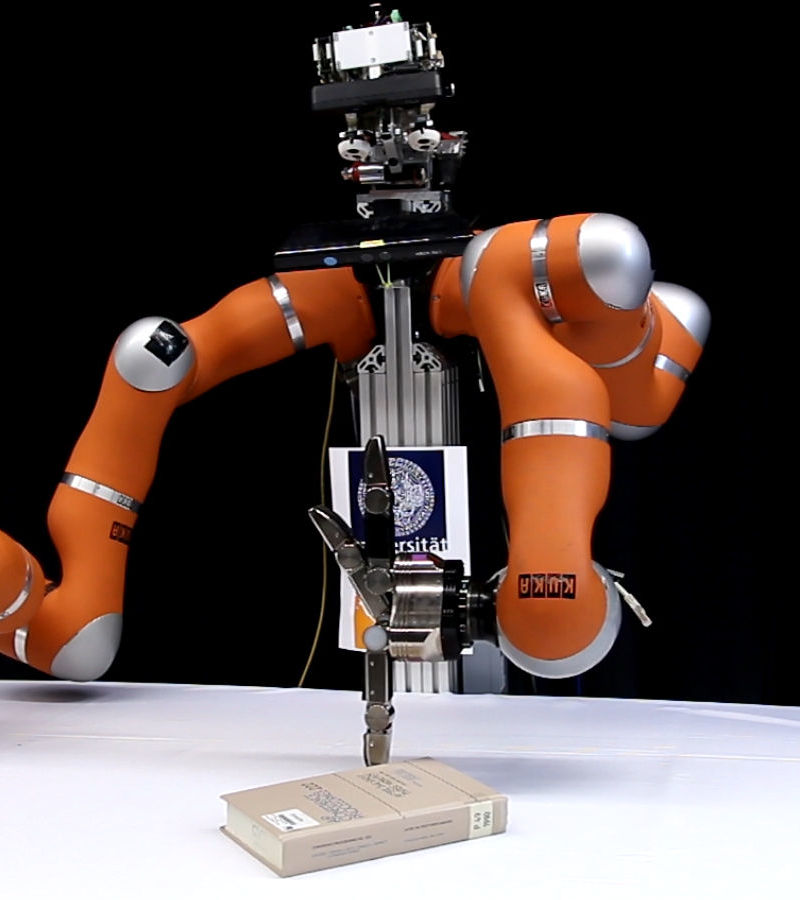}}
  \subfigure[\label{fig:pick3}{}]{\includegraphics[height=.18\linewidth]{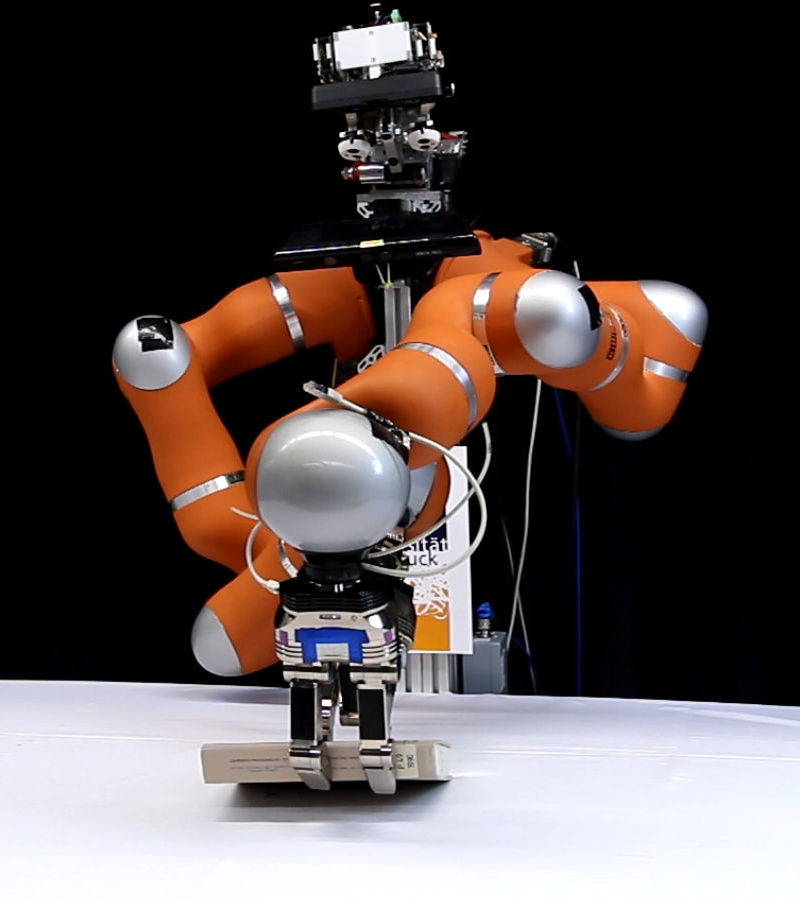}}
  \subfigure[\label{fig:pick4}{}]{\includegraphics[height=.18\linewidth]{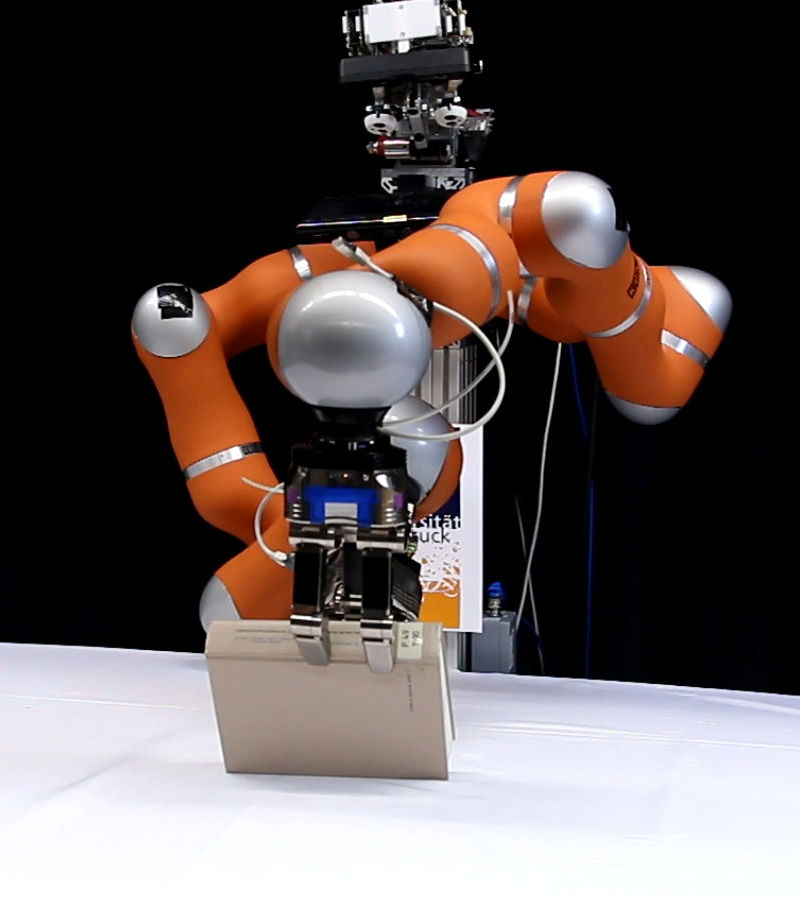}}
\caption{Robot setting (Fig. \ref{fig:robotsetting}) used for the experiments; Figs. \ref{fig:pick1} to \ref{fig:pick4} show a sample execution of the book grasping skill. In Fig. \ref{fig:pick1} the
sliding sensing action is visualized; Afterwards, the book is rotated by 90 degrees (Fig. \ref{fig:pick2}) and
lifted (Figs. \ref{fig:pick3}, \ref{fig:pick4}).}
\end{figure*}
In order to learn which sensing actions are discriminative for a given task and which
preparatory skills should be used in an observed state, PS provides a way to update
the transition weights by using external rewards.
The update is done with a modified version of the original
PS update rules \cite{Briegel2012}. Random-walk paths should be more
likely in future situations if the action taken was rewarded, i.e.\ the complex action
succeeded after performing the preparation action, and should be less likely otherwise.
Let $\{ s = c_{\text{1}} \rightarrow c_{\text{2}} \rightarrow \dots \rightarrow c_{\text{K}} = a \}$
be a random walk path that received a reward $\lambda^{(t)} \in \mathbb{R}$.
The weights are updated by
\begin{equation}
\underbrace{h^{t + 1}_{ij}}_{\text{next weight}} = \text{max}(1.0, \underbrace{h^t_{ij}}_{\text{current weight}} - \underbrace{\gamma \left(h^t_{ij} - 1 \right)}_{\text{damping}} + \rho_{ij} \underbrace{\lambda^{(t)}}_{\text{reward}})
\label{equ:learn1}
\end{equation}
where $h^t_{ij} = h^{t}\left(c_i, c_j\right)$. $\rho_{ij}$ is 1 if the path consists of a transition $c_i \rightarrow c_j$ and 0 otherwise.
The forgetting factor $\gamma$ defines how quickly the model forgets
previously-achieved rewards for a given path. It should only be nonzero if the robot
is placed in an environment where the behaviour of the objects changes slowly over time
(e.g., the object can break and change its physical properties after manipulating it
for a few hours).
\subsubsection{Building skill hierarchies}
If complex skill $A$ can be executed with a certain confidence, it is added to
the ECM of another complex skill $B$ by connecting it to each clip in layer 2 with the initial weight $h_{\textrm{init}}$.
The robot then goes back to the playing phase for skill $B$. If $B$ already has a high confidence, the transition weights
to certain preparatory skills will be high compared to $h_{\textrm{init}}$ and the probability of exploring the new preparatory skill $A$
is low but nonzero. If the confidence is low, all weights will be low and the robot will
start exploring the novel preparatory skill. Thus, PS is well suited for constructing skill hierarchies.
\section{Evaluation}
\label{sec:eval}
For evaluation we apply our approach to a \emph{complex book grasping}
task in an autonomous playing scenario. In a placement
task a skill hierarchy using the grasping skill is learned.
We use statistics (e.g.\ experimental success rates of skills) of the book grasping scenario
to \emph{simulate the convergence behaviour} of the same setting in case more
preparatory skills are used. We further show that the same
sensing actions and preparatory skills can be used for the different problem of
placing an object into a (closed) box.
\subsection{Applicability to real-world tasks}
\label{fig:evalbookpick}
We apply our method to a book grasping task (see video \footnote{\url{https://iis.uibk.ac.at/public/shangl/iros2016/iros.mpg}}).
The main challenge is to get a finger underneath the book in order to grasp it. The book is grasped by squeezing
it between both hands, lifting it on the binding side and then using in-hand manipulation to wrap the fingers around it.
It is easy to teach this skill for a single specific situation (e.g.\ by kinesthetic
teaching) but hard to generalize to arbitrary situations because of the complex interplay of
two arms and the book in several different orientations.

The experiments were performed with two KUKA LWR 4+ robotic arms with Schunk SDH grippers (Fig. \ref{fig:robotsetting}).
Different types of books (soft-cover, hard-cover, varying sizes) were used to train the haptic database.
All experiments were performed with the built-in impedance mode of the KUKA arms, which allows
books of different dimensions to be handled without explicitly coding them into the skills.
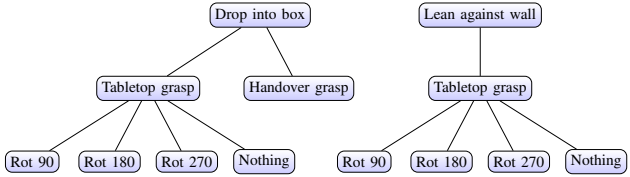
\begin{figure}[t!]
\centering
\scalebox{0.65} {
\begin{tikzpicture}[sibling distance = 5em,
  every node/.style = {shape=rectangle, rounded corners,
    draw, align = center,
    top color = white, bottom color = blue!20}, auto]]
    
    \begin{scope}
  \node{Drop into box}
    child {
    		node [xshift=-15mm] {Tabletop grasp}
   		child { node{Rot 90} }
   		child { node{Rot 180} }
   		child { node{Rot 270} }
   		child { node{Nothing} }
    }
    child{ node {Handover grasp} };
    \end{scope}

\begin{scope}[xshift=4.5cm]
  \node{Lean against wall}
    child {
    		node{Tabletop grasp}
   		child { node{Rot 90} }
   		child { node{Rot 180} }
   		child { node{Rot 270} }
   		child { node{Nothing} }
    };
\end{scope}
\end{tikzpicture}
}
\caption{Skill hierarchies for given complex skills (sub-skills with very low usage probability are omitted). For leaning vertically
the correct orientation of the book matters. For the tabletop grasp the book is always grasped at the binding side, whereas in the
handover grasp this is not the case (handover grasp is omitted). For dropping the book orientation does not matter and the hand-over grasp is considered.}
\label{fig:hierarchies}
\end{figure}

Three different sensing actions were used: Sliding (finger slides along the edge that is parallel to the table edges, while the second
hand keeps the object in place), Poking (the object is poked from the top) and Pressing (the book is squeezed between the 2 hands).
As preparatory skills we used a discretised version (90, 180 and 270 degrees) of a rotation controller that rotates the
object by an arbitrary angle. We also used a flipping controller (flipping the book upside down).
The reward was estimated automatically by measuring the force on the end-effector.
After rewarding the book was dropped onto the table and a random rotation action was
selected to prepare another random starting state. The learning parameters of the PS model were
set to $\lambda_{\textrm{succ}} = 1000$ (successful roll-outs), $\lambda_{\textrm{fail}} = -30$,
$h_{\textrm{init}} = 200$ and $\gamma = 0$ (no forgetting).

One of the challenges was to design robust controllers for autonomous play.
All the controllers and machine learning techniques were developed
within the \emph{kukadu} framework\footnote{\url{https://github.com/shangl/kukadu}}%
\footnote{\url{https://github.com/shangl/iros2016}}. The code is available
online and free to use. The robot was made to play for 100 roll-outs.
For creation of the haptic database the book was pushed clockwise by 90 degrees in order prepare the perceptual states autonomously.
At each rotation, 50 samples per sensing controller were collected.
A qualitative sketch of the learned ECM is shown in Fig.~\ref{fig:pickQualRes}.
The thick, coloured lines correspond to transitions with high weights and high probability.
The dominant sensing action is the sliding action, and its child
states have a semantic meaning, i.e., the orientation of the book. The transitions between the
state clips and the preparatory skills match the ground truth.
The execution sequence for a specific perceptual state after the playing phase is shown in Figs.\ \ref{fig:pick2}--\ref{fig:pick4}.
\begin{figure}[t!]
  \centering
  \subfigure[\label{fig:simconvspeca}{\footnotesize{Evolution of average success rate for $N_p = 6$ (blue), $N_p = 20$ (red), $N_p = 30$ (green), $N_p = 39$ (yellow)}}]{\includegraphics[width=.46\linewidth]{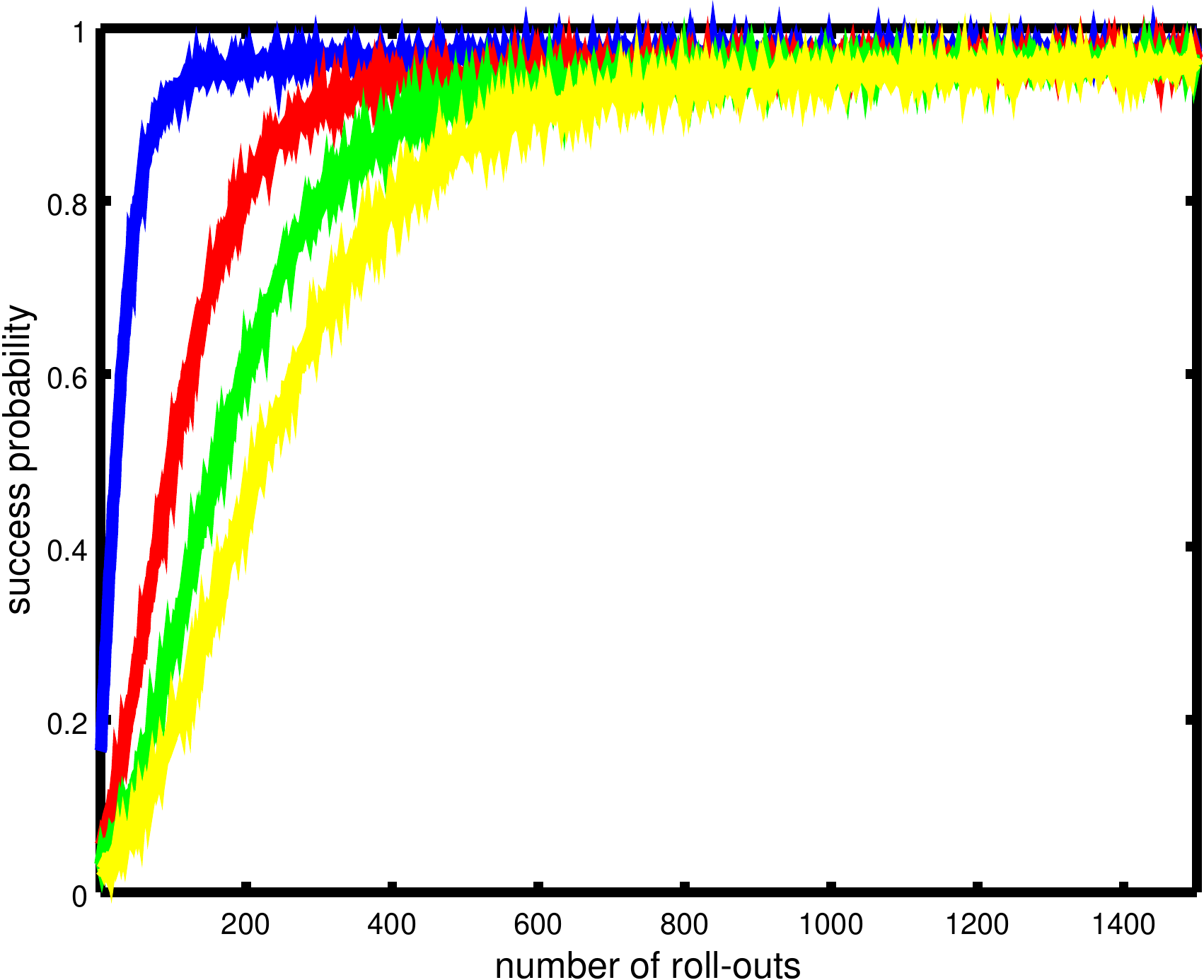}}\hfill
  \subfigure[\label{fig:simconvspecb}{\footnotesize{Number of roll-outs $N_r$ required to reach a success rate of 0.9 for different $N_p$.}}]{\includegraphics[width=.46\linewidth]{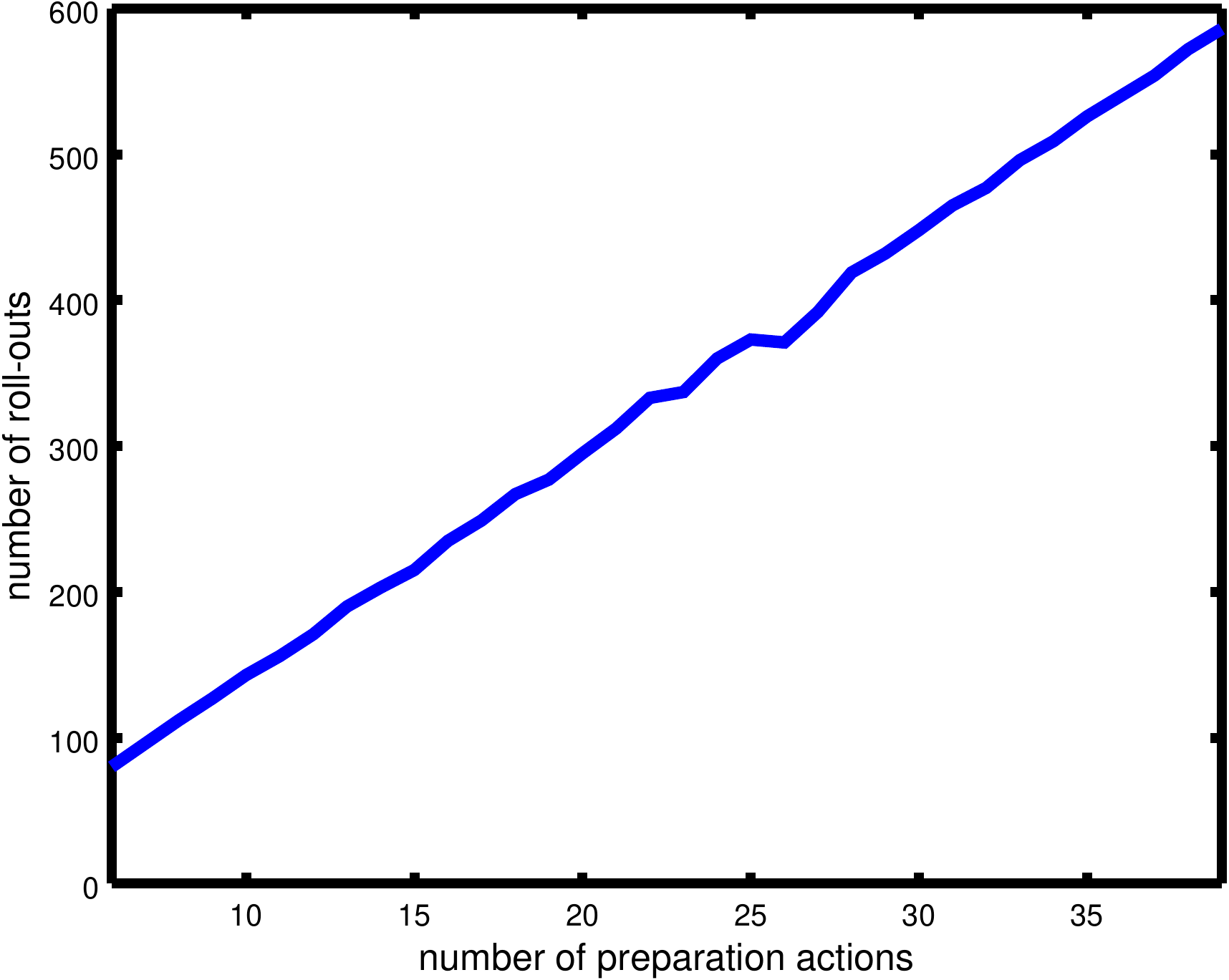}}
  \caption{Convergence behaviour of the proposed hierarchical skill learning model}
  \label{fig:simconvspec}
\end{figure}
After training, the tabletop grasping strategy was added to the set of preparatory skills, and
two placement strategies (drop into box, lean against a wall) were shown by kinesthetic teaching. The robot 
was able to learn that it has to grasp the book first.
In that scenario, the robot constructed the skill hierarchies shown in Fig.~\ref{fig:hierarchies}.
\subsection{Convergence simulation}
An important property is the success rate convergence. For the
book picking task the ground truth is known, as the perceptual states of the sliding action are
semantically meaningful, i.e.\ the orientation of the book. Success rate statistics
for the implemented controllers were measured during the real experiment and were used
to simulate the convergence behaviour if some useless preparatory skills were added.
The probability of correct classification
was given by $p_{\textrm{slide}} = 0.93$ (sliding action), 
$p_{\textrm{poke}} = 0.27$ (poking action), $p_{\textrm{press}} = 0.4$
(pressing action). The success probability of the grasping action was $p_p = 0.98$ (given that the correct perceptual state was estimated).
As the used projective simulation model
is based on a stochastic process, $N = 10000$ agents
(each agent is a simulated separate robot) were executed over $t = 1500$ roll-outs for different numbers
of preparatory actions $N_p$. The success in each
time step was averaged over all agents (Fig. \ref{fig:simconvspeca}).
Fig. \ref{fig:simconvspecb} shows how many roll-outs were required to reach a success rate $p_{\textrm{succ}} = 0.9$
for different $N_p$. For the special case of $N_p = 6$ the number of roll-outs was determined with $N_r = 80$. A trivial
system that tries every combination of $N_s = 3$ sensing actions, $N_o = 4$ sensing outcomes per action and $N_p = 6$ preparation
actions would require $N_r = N_s N_o N_p = 72$ roll-outs to observe every combination only once. However, from this information
it is difficult to infer knowledge about a certain sensing / preparation skill combination (e.g., success / failure could be caused by
noise, unexpected temporary circumstances, etc.). Our method only requires 80 roll-outs and focusses
on regions in the exploration space that are interesting for the problem and is able to handle such kind of noise.
\subsection{Task diversity}
\begin{figure*}
\centering
  \subfigure[\label{fig:book1}{}]{\includegraphics[width=.15\linewidth]{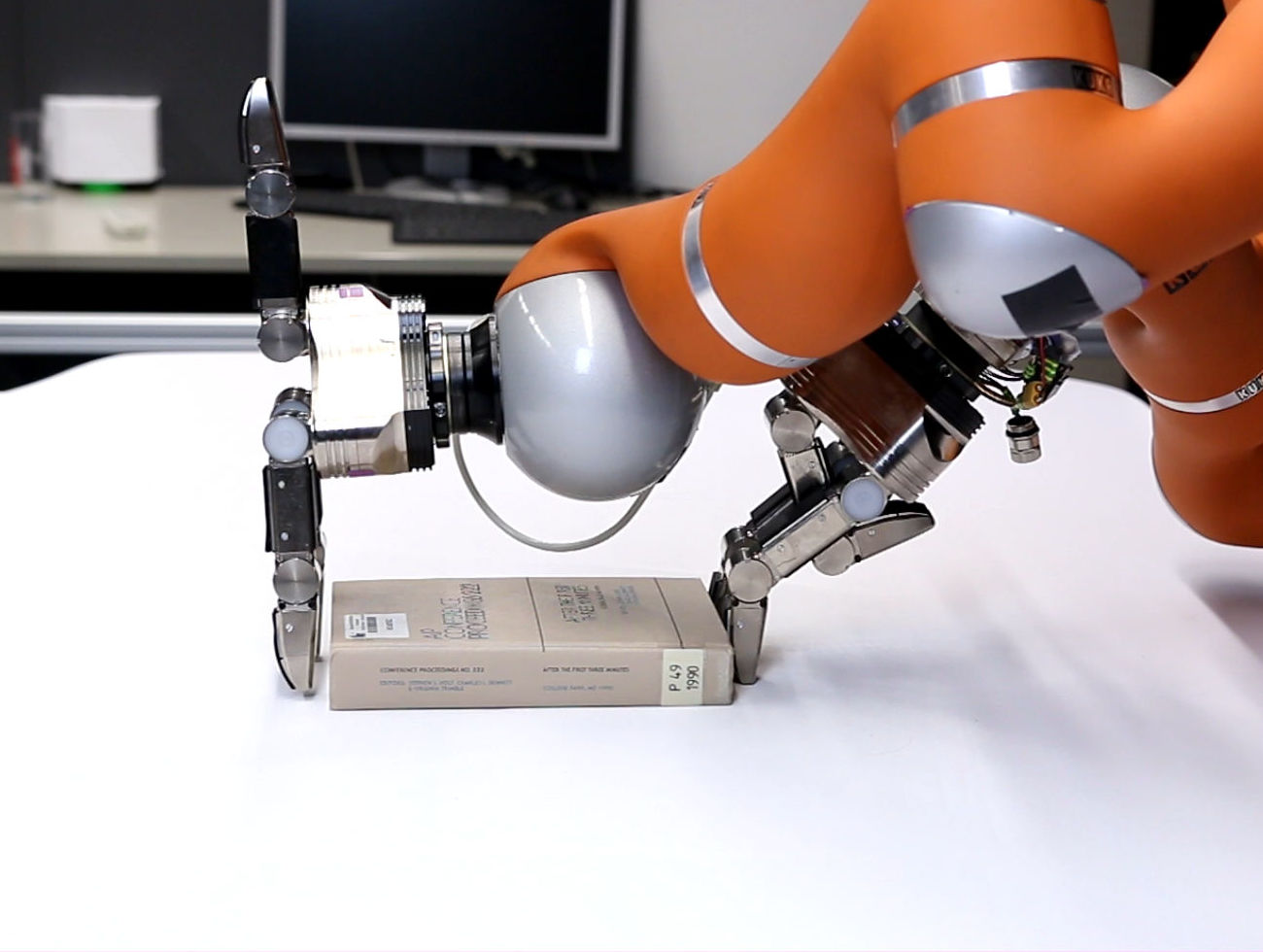}}
  \subfigure[\label{fig:book2}{}]{\includegraphics[width=.15\linewidth]{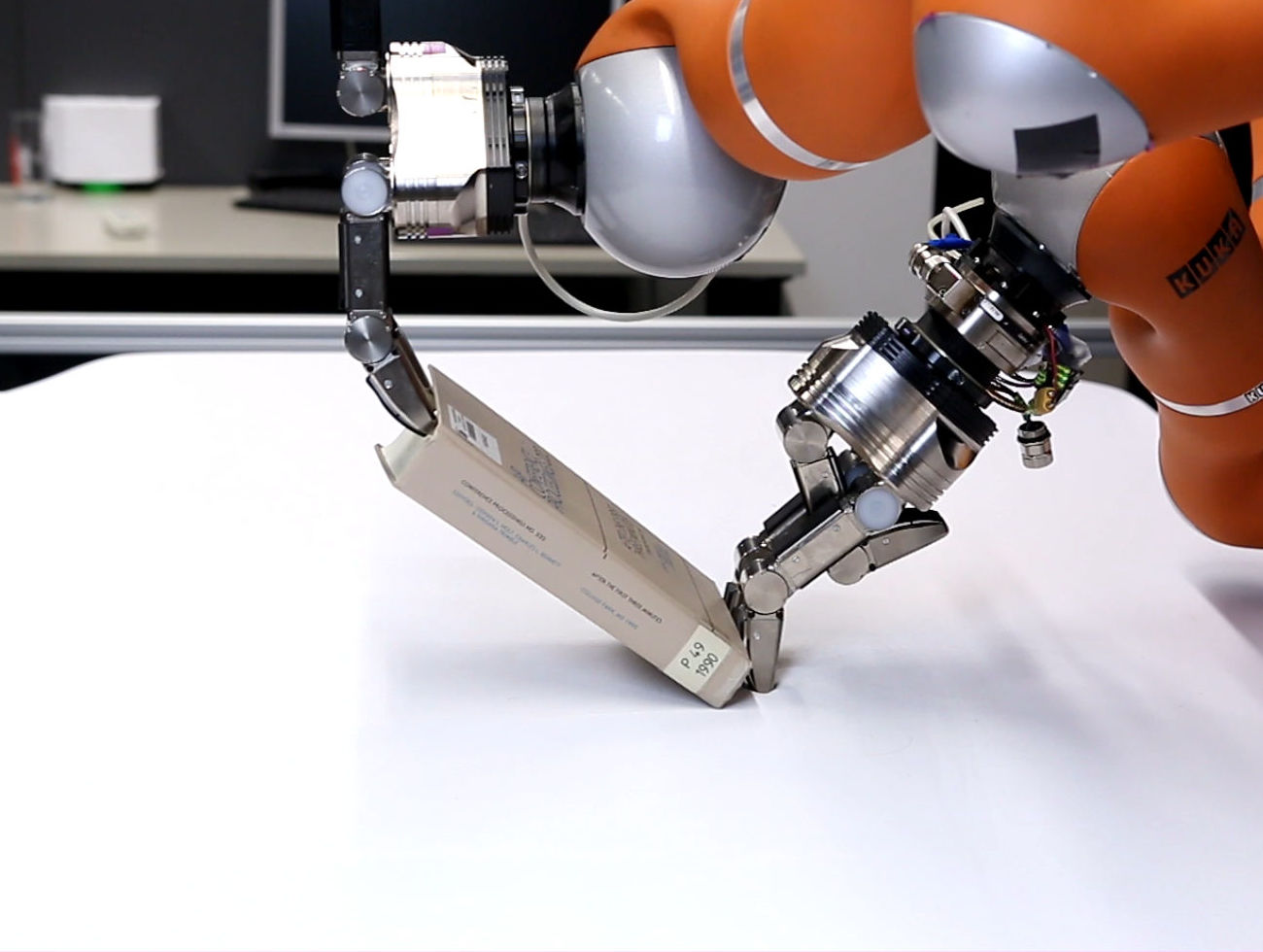}}
  \subfigure[\label{fig:book3}{}]{\includegraphics[width=.15\linewidth]{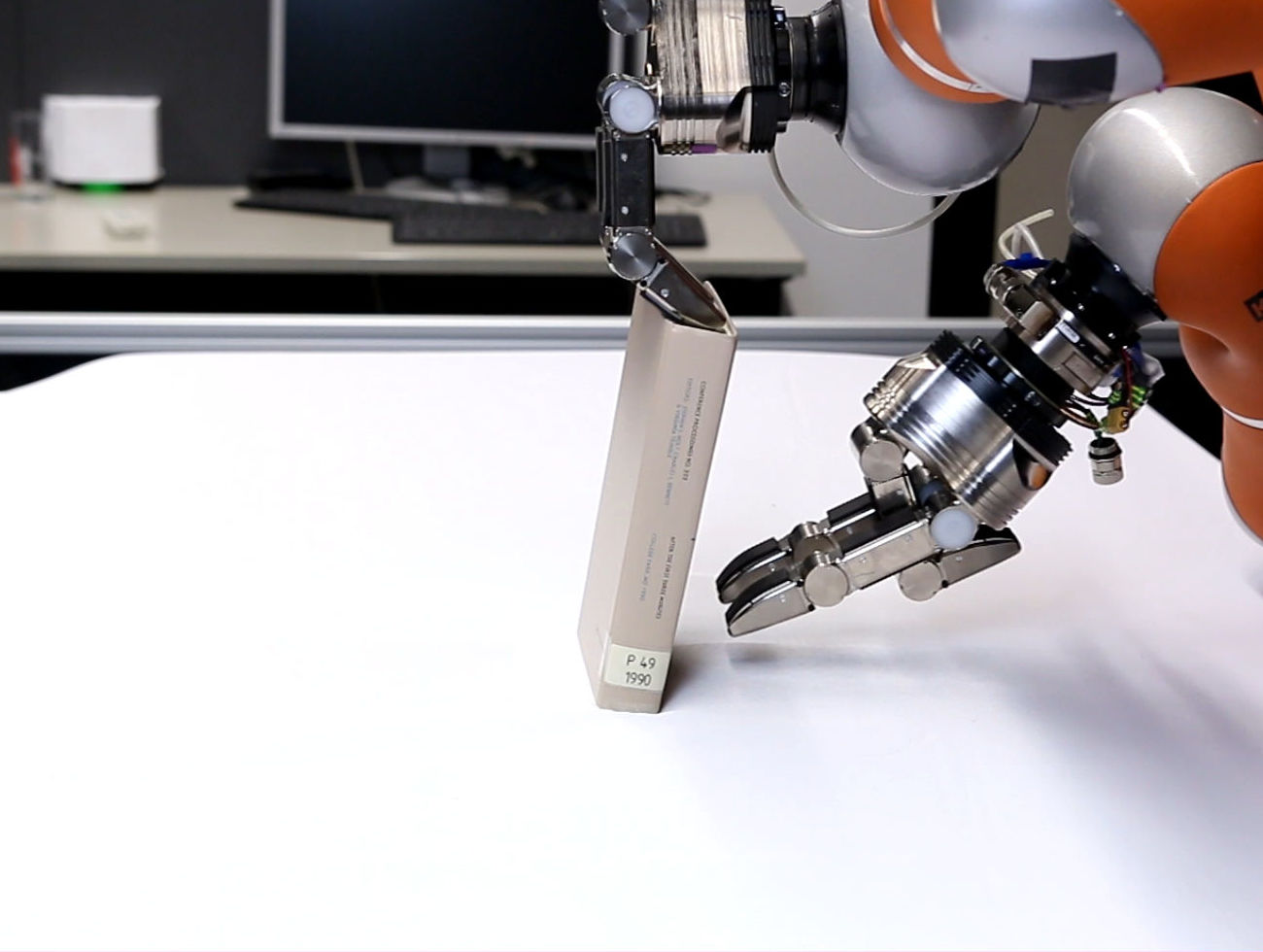}}
  \subfigure[\label{fig:book4}{}]{\includegraphics[width=.15\linewidth]{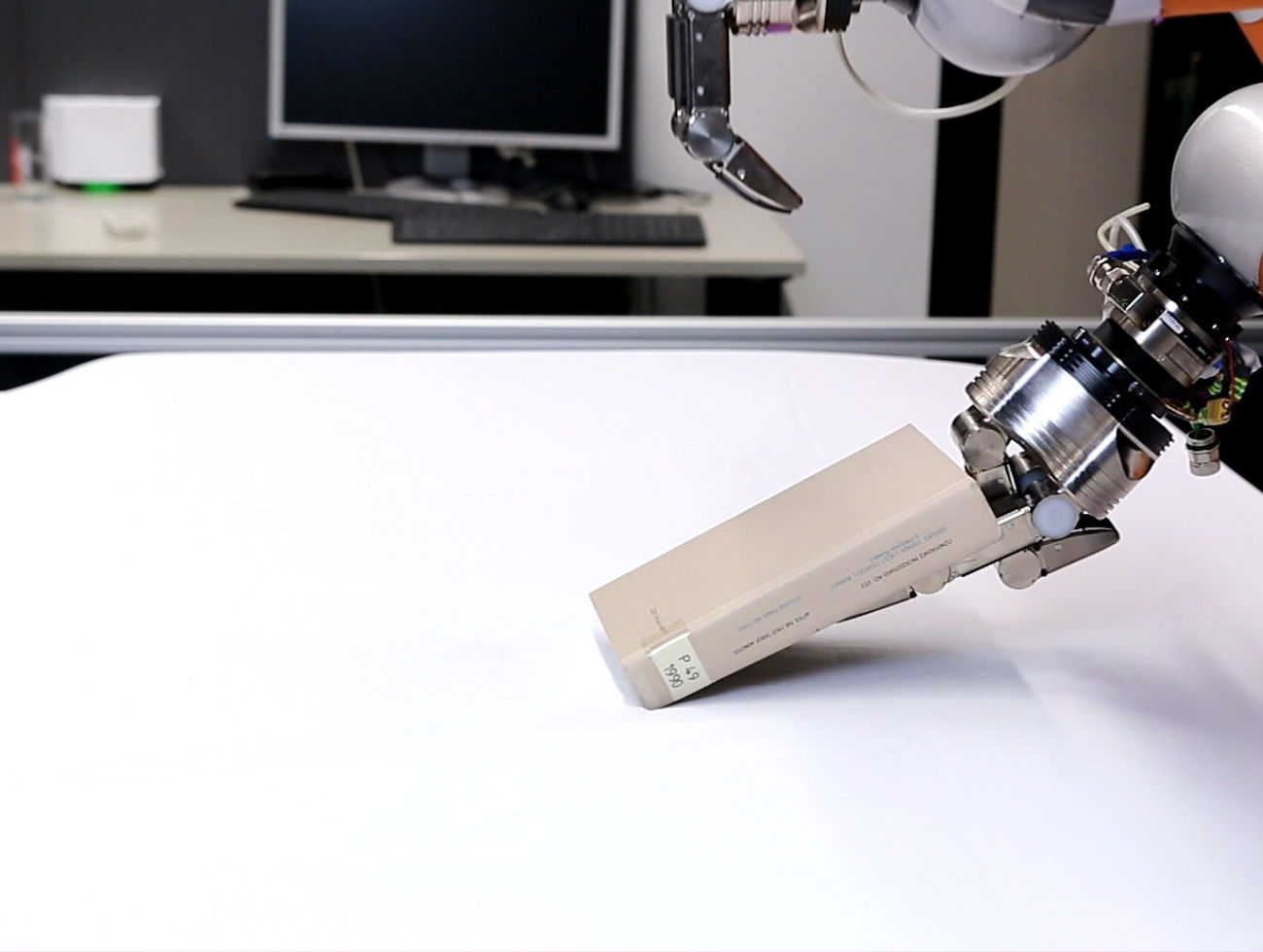}}
  \subfigure[\label{fig:book5}{}]{\includegraphics[width=.15\linewidth]{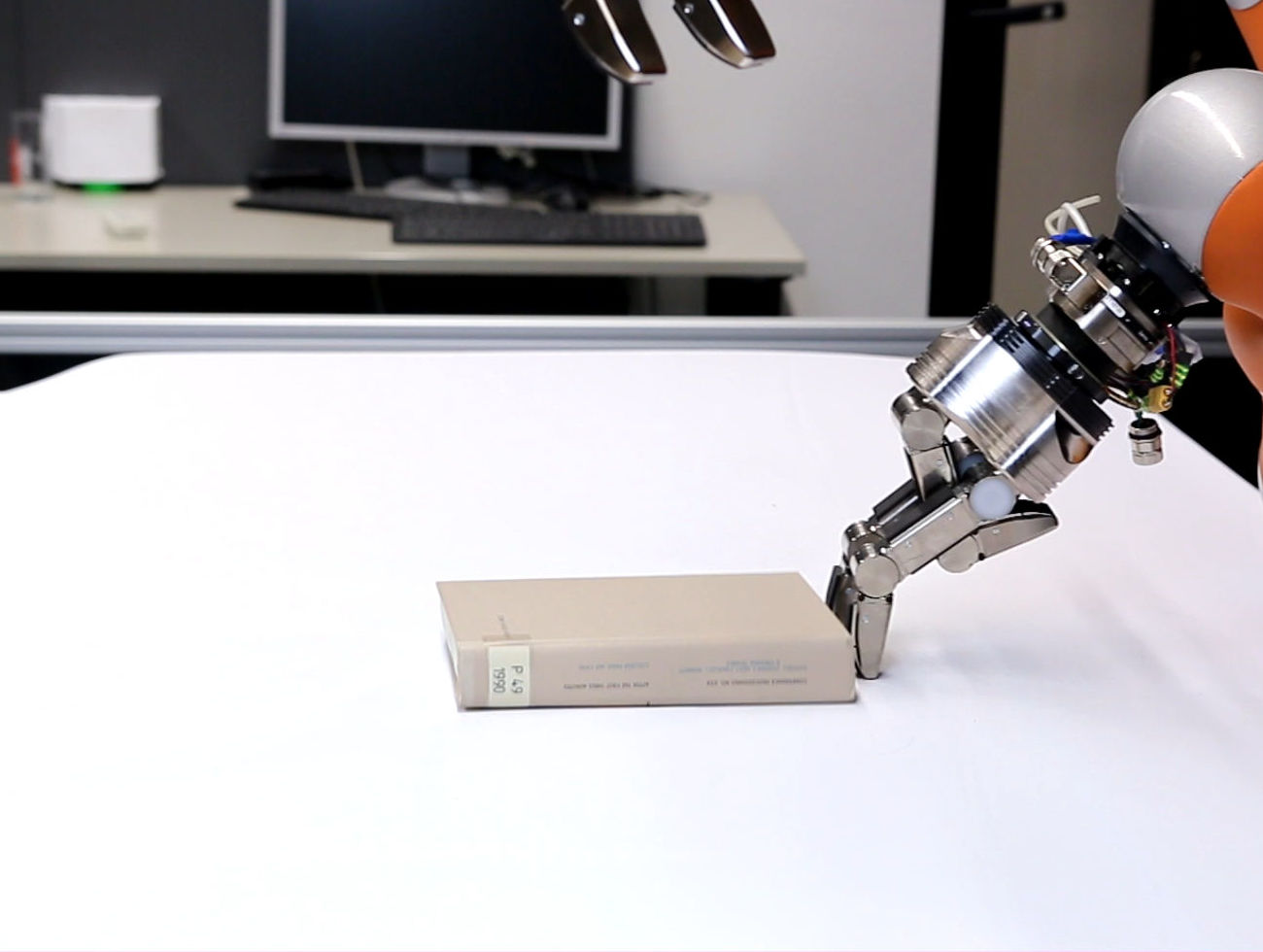}}
  \subfigure[\label{fig:book6}{}]{\includegraphics[width=.15\linewidth]{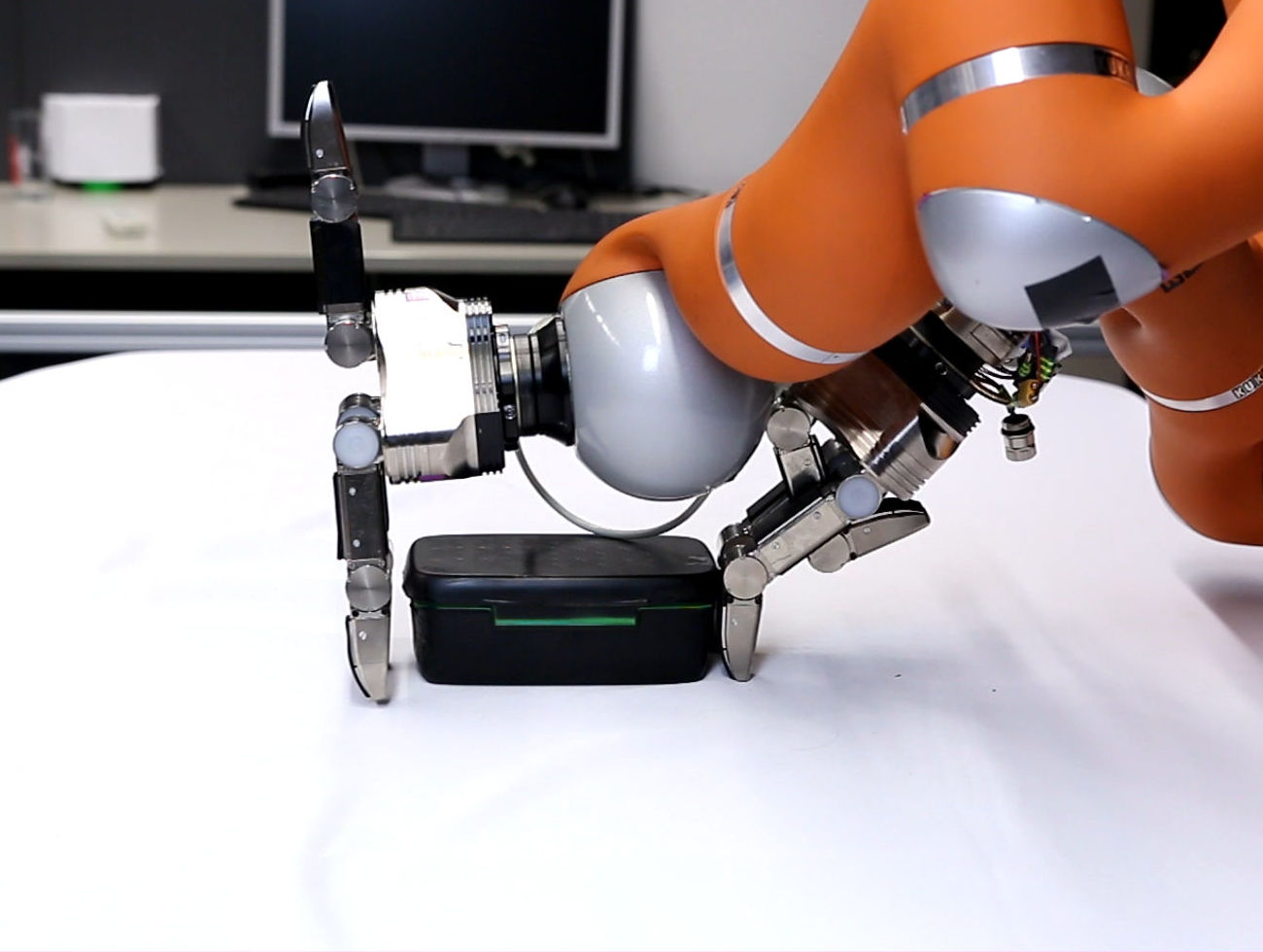}}
  \subfigure[\label{fig:book7}{}]{\includegraphics[width=.15\linewidth]{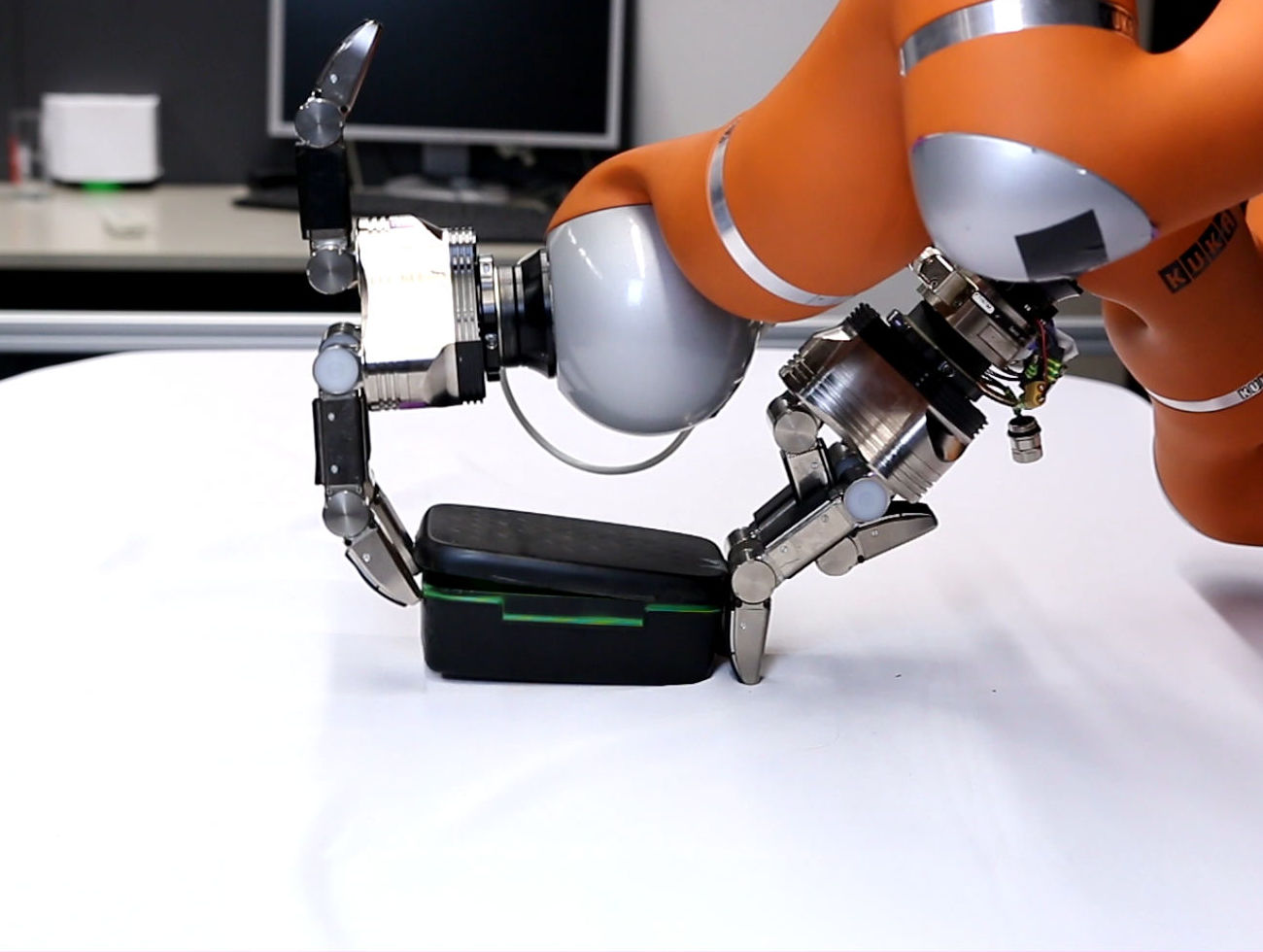}}
  \subfigure[\label{fig:book8}{}]{\includegraphics[width=.15\linewidth]{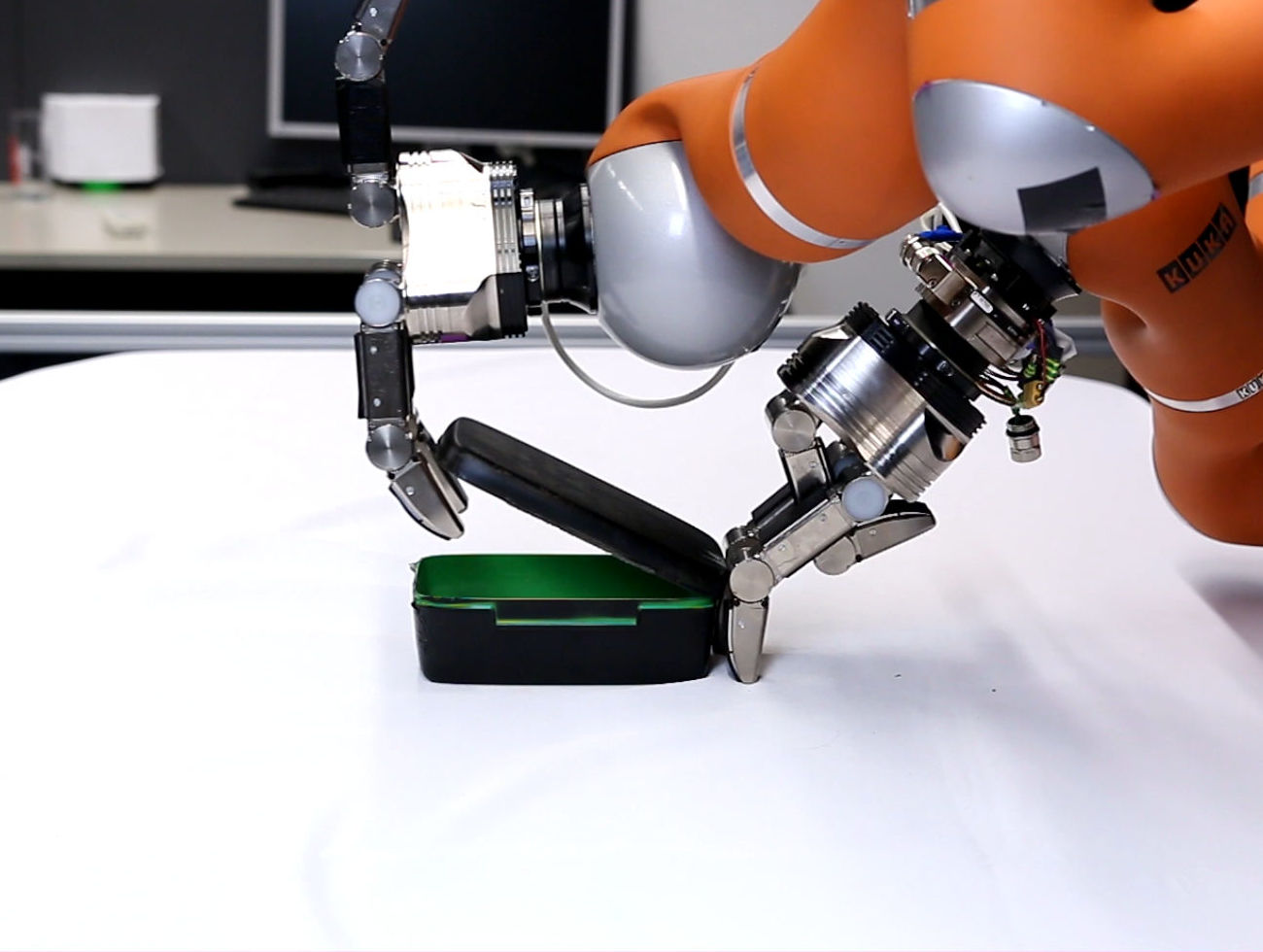}}
  \subfigure[\label{fig:book9}{}]{\includegraphics[width=.15\linewidth]{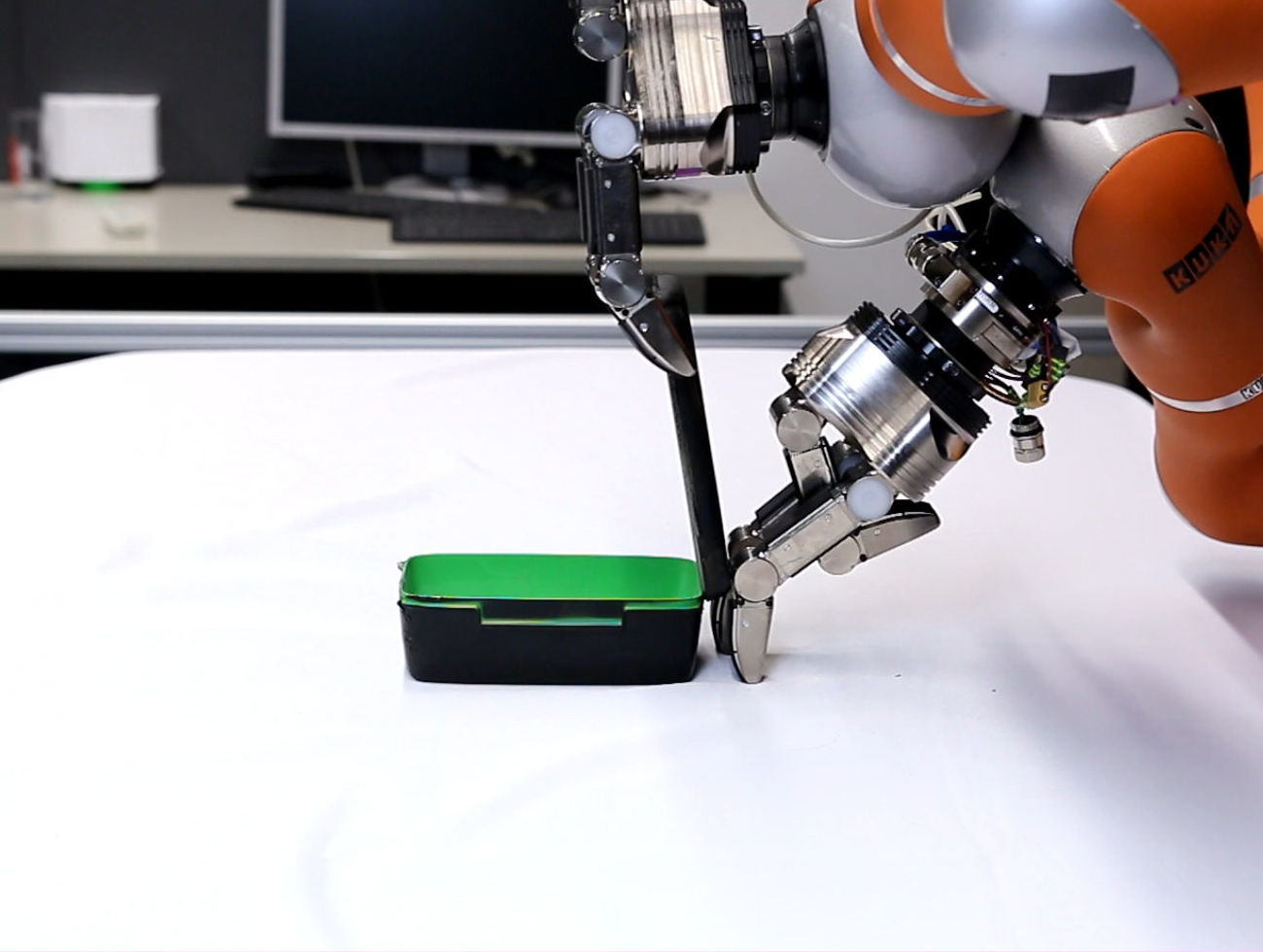}}
  \subfigure[\label{fig:book10}{}]{\includegraphics[width=.15\linewidth]{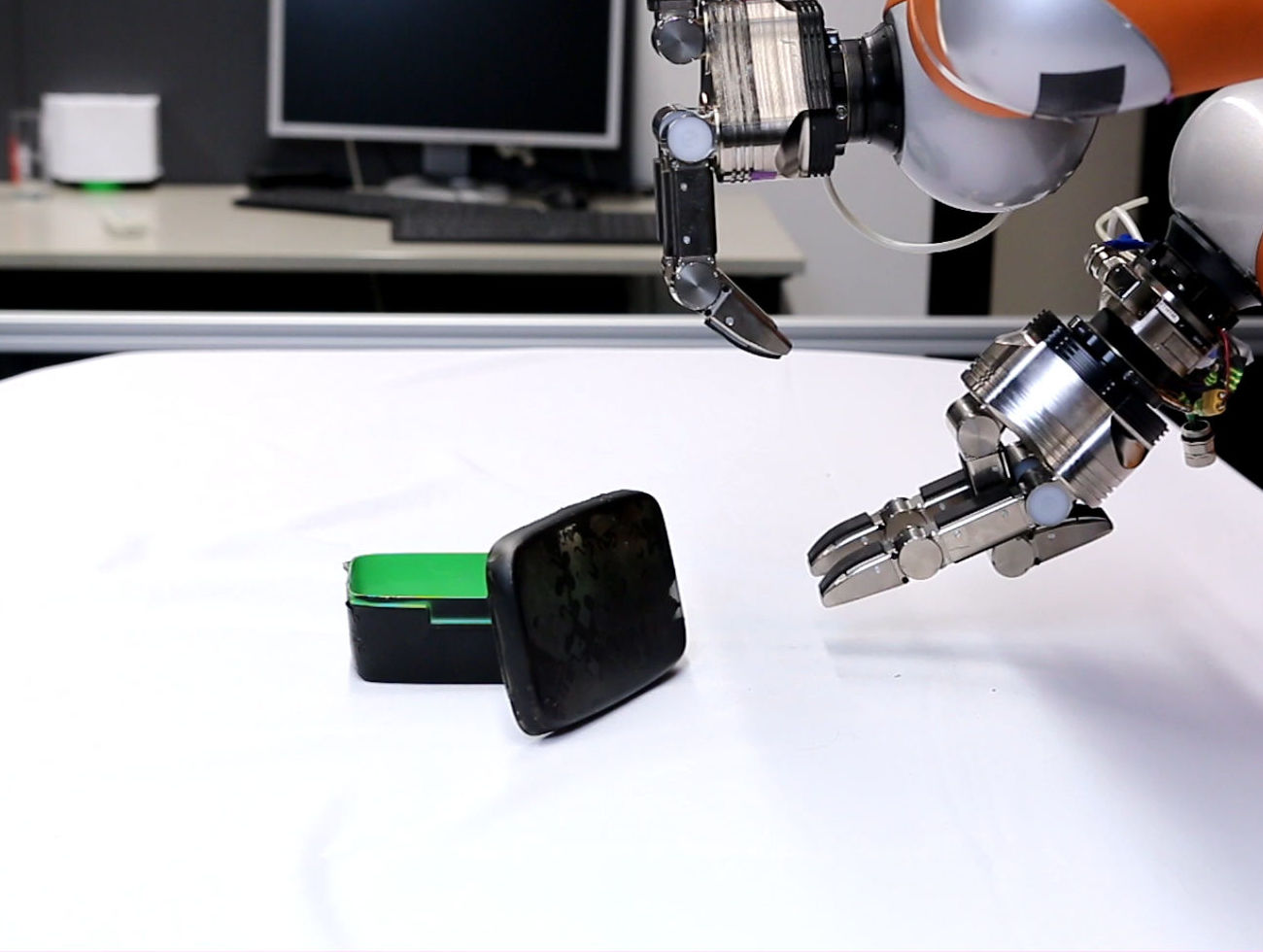}}
  \caption{Flipping skill performed in two different tasks -- providing two different semantic effects. In the book picking
  scenario, the book got flipped. In the box opening scenario, the box got opened.}
  \label{fig:bookboxcomp}
\end{figure*}
In Section \ref{fig:evalbookpick}, a specific example for the applicability of the method was shown. We now
show that the same setting can be used to learn a diverse set of skills.

In the book grasping scenario the orientation of the book was estimated by sliding. However,
our method does not assume such semantic categories and is not
bound to any specific property (e.g.\ pose) of the object. To illustrate this, the system with the same
sensing and preparatory controllers was confronted with a small rectangular food-box with a removable cover
(Fig. \ref{fig:robotsetting}). The task was to place an object inside the box. By kinesthetic teaching
the robot was taught to grasp an object and place it inside the open box. For the generation of the haptic
database, the box was placed in front of the robot in the open and closed configurations.
Poking was determined to be the best sensing action. From a human perspective it is reasonable to poke
the top of a box to determine whether it is open or not.
The flipping skill of the book picking scenario turned out
to be able to remove the cover from the box as a preparatory action to place something inside the box
(for comparison see Fig.~\ref{fig:bookboxcomp}).

We emphasize that the robot does not have any notion of the semantic meaning of the actions it performed.
Still it came up with a semantically meaningful selection of skills and was able to achieve the task
with the same skill set as in the book picking scenario.
\subsection{Limitations}
The design choices responsible for the high degree of autonomy and fast convergence,
of course, come with some disadvantages and limitations. The strongest restriction
comes from the open-loop nature. No effects are predicted, and in the current version there is no explicit
way to check whether the task is still going well during execution. However, implicit error handling
mechanisms can be hidden inside the preparatory controllers, which puts high demands on them.
Further, complex planning outside of the skill hierarchies is not possible as there is no environment
model. This is a disadvantage in case of multi-object tasks. Another restriction arises
from the loose coupling of controllers in very dynamic tasks. For example, a ball might roll away
between preparation and execution of the complex skill.
\section{Conclusions}
We introduced a novel method for robotic object manipulation. The key idea is to teach novel skills
achieving complex tasks with limited generalisation capabilities. Previously-trained skills are then
used to prepare the environment such that the limited controllers can still succeed. This way,
the teaching of novel skills is simple. In this framework, learning can be mostly autonomous.
Further, complex skill hierarchies can be constructed by adding learned complex skills to the set
of preparatory skills.

The approach was evaluated in a complex pick-and-place task, where a book can be placed in
several different ways. The system was able to learn that it has to grasp the book before
it can be placed. Further it learns that the book has to be rotated to the right orientation such
that it can be grasped from a table with a complex sequence of in-hand manipulations.
Additionally, the same set of sensing and preparatory skills can be applied to a wide set of
different problems. This was shown by using the same preparatory skill set for putting an object inside a closed box.
It succeeded to open the box by the flipping action in order to place an object inside.
Further, the convergence of the learning approach was evaluated in simulation when more preparatory actions are used.
It was found that during the exploration phase, the robot focusses on regions that are
of particular interest to the problem. Therefore it can learn a skill with high confidence
with just a slightly higher number of roll-outs compared to executing every combination
only once.



\section*{Acknowledgment}
The research leading to these results has received funding
from the European Communitys Seventh Framework Programme FP7/20072013 (Specific Programme Cooperation,
Theme 3, Information and Communication Technologies) under grant agreement no. 610532, Squirrel.
Special thanks to Hans J. Briegel and Vedran Dunjko for the very helpful and fruitful
discussions and guidance.


\bibliographystyle{IEEEtran}
\bibliography{bibfile}

\end{document}